\documentclass{article}

\usepackage{microtype}
\usepackage{graphicx}
\usepackage{subcaption}
\usepackage{booktabs} 
\usepackage{multirow}

\usepackage{hyperref}

\usepackage[preprint]{icml2026}

\usepackage{amsmath}
\usepackage{amssymb}
\usepackage{mathtools}
\usepackage{amsthm}

\usepackage{amsmath,amsfonts,bm}

\def\eqref#1{equation~\ref{#1}}

\def\1{\bm{1}}

\DeclareMathAlphabet{\mathsfit}{\encodingdefault}{\sfdefault}{m}{sl}
\SetMathAlphabet{\mathsfit}{bold}{\encodingdefault}{\sfdefault}{bx}{n}

\usepackage{calc}
\usepackage{placeins}
\usepackage{tikz}
\usetikzlibrary{decorations.pathreplacing,positioning, calc, patterns, patterns.meta}
\usepackage[export]{adjustbox}

\usepackage[capitalize,noabbrev]{cleveref}

\definecolor{dpurple}{HTML}{884EA0}
\definecolor{dblue}{HTML}{FF0000}
\definecolor{dgreen}{HTML}{00A08A}
\definecolor{dyellow}{HTML}{F2AD00}
\definecolor{dgray}{HTML}{5BBCD6}

\newsavebox{\puppybox}
\newlength{\puppywidth}
\newlength{\puppyheight}
\newlength{\patchtrimleft}
\newlength{\patchtrimright}
\newlength{\patchtrimtop}
\newlength{\patchtrimbottom}

\newcommand{\ablationoriginal}[2][0.12\textwidth]{%
  \includegraphics[width=#1]{figures/token_ablation_images/image#2/original.png}%
}
\newcommand{\ablationtoken}[3][0.12\textwidth]{%
  \includegraphics[width=#1]{figures/token_ablation_images/image#2/token_#3.png}%
}

\icmltitlerunning{Laminating Representation Autoencoders for Efficient Diffusion}

\begin{document}

\twocolumn[
  \icmltitle{Laminating Representation Autoencoders for Efficient Diffusion}


  \begin{icmlauthorlist}
    \icmlauthor{Ram\'on Calvo-Gonz\'alez}{unige}
    \icmlauthor{Fran\c{c}ois Fleuret}{unige,fair}
  \end{icmlauthorlist}

  \icmlaffiliation{unige}{Department of Computer Science, University of Geneva, Geneva, Switzerland}
  \icmlaffiliation{fair}{FAIR, Meta}

  \icmlcorrespondingauthor{Ram\'on Calvo-Gonz\'alez}{ramon.calvogonzalez@unige.ch}

  \icmlkeywords{Self-Supervised Learning, World Models, Vision Transformers, Representation Learning}

  \vskip 0.3in
]

\printAffiliationsAndNotice{}

\begin{abstract}
Recent work has shown that diffusion models can generate high-quality images by operating directly on SSL patch features rather than pixel-space latents. However, the dense patch grids from encoders like DINOv2 contain significant redundancy, making diffusion needlessly expensive. We introduce FlatDINO, a variational autoencoder that compresses this representation into a one-dimensional sequence of just 32 continuous tokens---an 8$\times$ reduction in sequence length and 48$\times$ compression in total dimensionality. On ImageNet 256$\times$256, a DiT-XL trained on FlatDINO latents achieves a gFID of 1.85 with classifier-free guidance while requiring 8$\times$ fewer FLOPs per forward pass and up to 4.5$\times$ fewer FLOPs per training step compared to diffusion on uncompressed DINOv2 features. These are preliminary results and this work is in progress.
\end{abstract}

\section{Introduction}

\begin{figure}[!t]
\centering
\includegraphics[width=0.85\columnwidth]{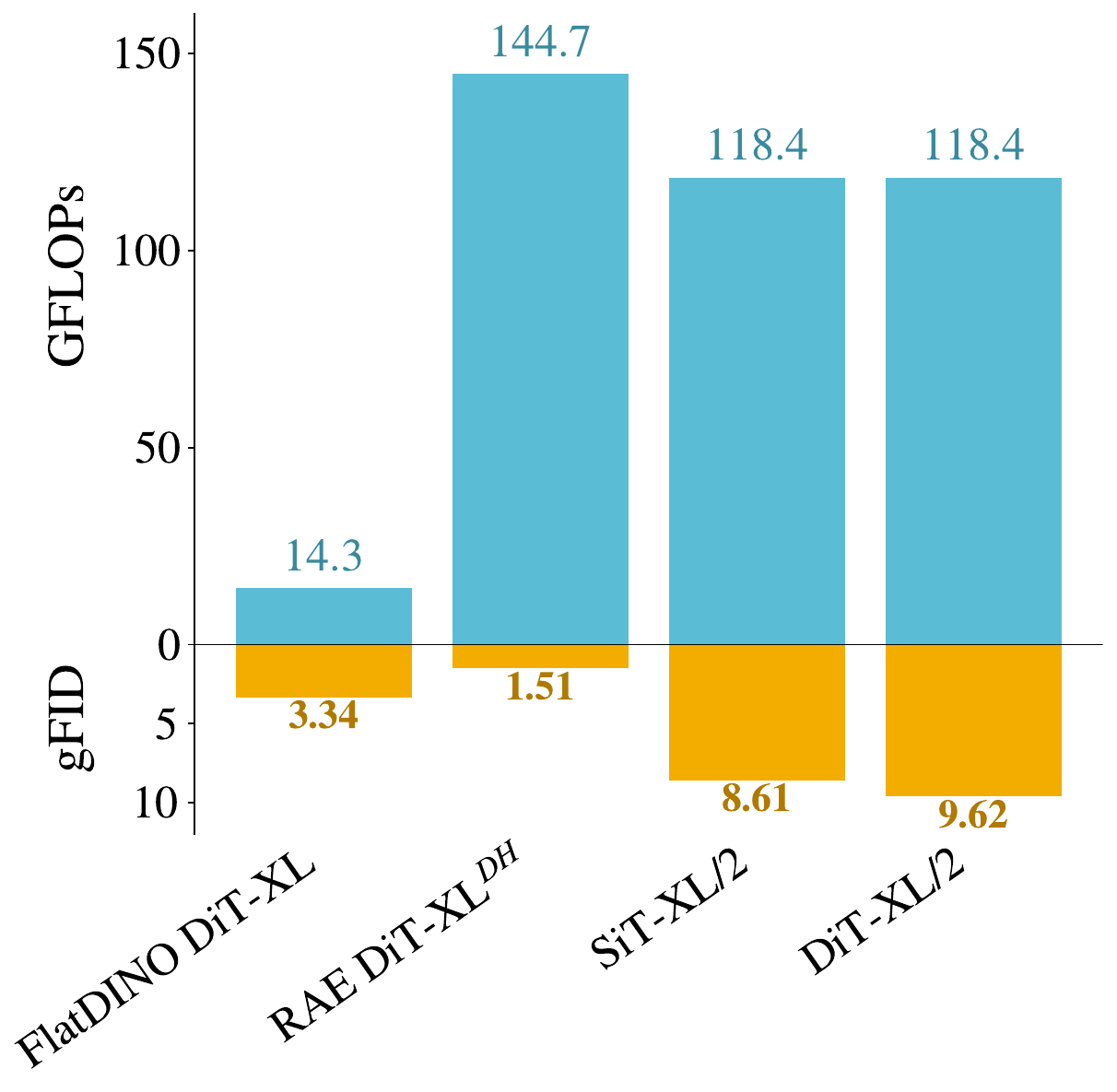}
\caption{GFLOPs per forward pass versus gFID (without CFG) for similar-sized diffusion transformers on ImageNet 256$\times$256. FlatDINO (ours) achieves a substantial reduction in FLOPs while maintaining competitive generation quality.}
\label{fig:flops-vs-fid}
\end{figure}

\begin{figure*}[!t]
\centering
\sbox{\puppybox}{\includegraphics{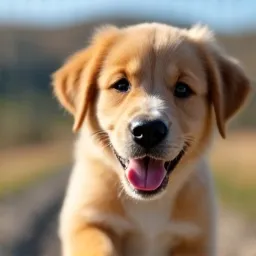}}
\setlength{\puppywidth}{\wd\puppybox}
\setlength{\puppyheight}{\ht\puppybox}
\addtolength{\puppyheight}{\dp\puppybox}
\begin{tikzpicture}[x=1cm, y=1cm, >=stealth]

    \def\ImageWidth{1.8cm}
    \def\ImageHeight{1.8cm}
    \node[inner sep=0pt] (image) at (0, -1.2) {\includegraphics[width=\ImageWidth, height=\ImageHeight]{figures/puppy.png}};
    \draw[white, line width=0.8pt] let \p1=(image.south west), \p2=(image.north east), \n1={0.45cm} in
        foreach \i in {1,...,3} {
            (\x1 + \i*\n1, \y1) -- (\x1 + \i*\n1, \y2)
        }
        foreach \i in {1,...,3} {
            (\x1, \y1 + \i*\n1) -- (\x2, \y1 + \i*\n1)
        };

    \draw[->, thick] (0.95, -1.2) -- (1.85, -1.2);

    \foreach \col in {0,1,2,3,4} {
        \pgfmathsetmacro{\ypos}{-0.4 - 0.4*\col}
        \pgfmathsetmacro{\patchcol}{mod(\col, 4)}
        \setlength{\patchtrimleft}{\dimexpr\patchcol\puppywidth/4\relax}
        \setlength{\patchtrimright}{\dimexpr3\puppywidth/4 - \patchcol\puppywidth/4\relax}
        \setlength{\patchtrimtop}{0pt}
        \setlength{\patchtrimbottom}{\dimexpr3\puppyheight/4\relax}
        \node[anchor=center, inner sep=0pt] at (2.15, \ypos) {\adjincludegraphics[width=0.3cm, trim={{\the\patchtrimleft} {\the\patchtrimbottom} {\the\patchtrimright} {\the\patchtrimtop}}, clip]{figures/puppy.png}};
        \draw[line width=0.5pt] (2.0, \ypos-0.15) rectangle (2.3, \ypos+0.15);
    }
    \node[font=\scriptsize] at (2.15, -2.3) {$\vdots$};

    \draw[line width=1pt] (2.4, -2.5) rectangle (4.4, 1.0);
    \node[align=center] at (3.4, -0.75) {DINOv2\\ViT-B/14\\w/ Reg.};

    \draw[fill=dblue] (4.5, 0.65) rectangle (4.8, 0.95);
    \foreach \i in {0, 1} {
        \pgfmathsetmacro{\ypos}{0.65 - 0.4*(\i+1)}
        \draw[fill=dpurple] (4.5, \ypos) rectangle (4.8, \ypos+0.3);
    }
    \foreach \i in {0, 1, 2, 3, 4} {
        \pgfmathsetmacro{\ypos}{0.65 - 0.4*(\i+3)}
        \draw[fill=dgray] (4.5, \ypos) rectangle (4.8, \ypos+0.3);
    }
    \node[font=\scriptsize] at (4.65, -2.3) {$\vdots$};

    \foreach \i in {0, 1, 2} {
        \pgfmathsetmacro{\ypos}{0.65 - 0.4*(\i)}
        \draw[pattern={Lines[angle=45, distance=2pt, line width=1pt]}, pattern color=dgreen] (5.2, \ypos) rectangle (5.5, \ypos+0.3);
    }
    \foreach \i in {0, 1, 2, 3, 4} {
        \pgfmathsetmacro{\ypos}{0.65 - 0.4*(\i+3)}
        \draw[fill=dgray] (5.2, \ypos) rectangle (5.5, \ypos+0.3);
    }
    \node[font=\scriptsize] at (5.35, -2.3) {$\vdots$};

    \draw[line width=1pt] (5.6, -2.5) rectangle (7.2, 1.0);
    \node[align=center] at (6.4, -0.75) {ViT\\Encoder};

    \foreach \i in {0, 1, 2} {
        \pgfmathsetmacro{\ypos}{0.65 - 0.4*(\i)}
        \draw[fill=dgreen] (7.3, \ypos) rectangle (7.9, \ypos+0.3);
        \draw[dashed] (7.6, \ypos) -- (7.6, \ypos+0.3);
    }
    \node[font=\scriptsize] at (7.6, -0.35) {$D_{\mathrm{KL}}$};

    \draw[->, thick] (7.95, 0.4) -- (8.55, 0.4);

    \foreach \i in {0, 1, 2} {
        \pgfmathsetmacro{\ypos}{0.65 - 0.4*(\i)}
        \draw[fill=dgreen] (8.6, \ypos) rectangle (8.9, \ypos+0.3);
    }
    \foreach \i in {0, 1, 2, 3, 4} {
        \pgfmathsetmacro{\ypos}{0.65 - 0.4*(\i+3)}
        \draw[pattern={Lines[angle=45, distance=2pt, line width=1pt]}, pattern color=dyellow] (8.6, \ypos) rectangle (8.9, \ypos+0.3);
    }
    \node[font=\scriptsize] at (8.75, -2.3) {$\vdots$};
    \draw[line width=1pt] (9.0, -2.5) rectangle (10.6, 1.0);
    \node[align=center] at (9.8, -0.75) {ViT\\Decoder};

    \foreach \i in {0, 1, 2} {
        \pgfmathsetmacro{\ypos}{0.65 - 0.4*(\i)}
        \draw[fill=dgreen] (10.7, \ypos) rectangle (11.0, \ypos+0.3);
    }
    \foreach \i in {0, 1, 2, 3, 4} {
        \pgfmathsetmacro{\ypos}{0.65 - 0.4*(\i+3)}
        \draw[fill=dyellow] (10.7, \ypos) rectangle (11.0, \ypos+0.3);
    }
    \node[font=\scriptsize] at (10.85, -2.3) {$\vdots$};

    \draw[->, dashed, thick, gray] (5.35, -2.7) -- (5.35, -3.2) -- (7.05, -3.2);
    \draw[->, dashed, thick, gray] (10.85, -2.7) -- (10.85, -3.2) -- (8.95, -3.2);
    \node[font=\scriptsize] at (8.0, -3.2) {$\mathcal{L}_{\text{MSE}}($\tikz[baseline=0.1ex]{\draw[fill=dgray] (0,0) rectangle (0.2,0.2);}$,$\tikz[baseline=0.1ex]{\draw[fill=dyellow] (0,0) rectangle (0.2,0.2);}$)$};

\end{tikzpicture}
  \caption{A frozen DINOv2 ViT-B/14 with registers \citep{darcetVisionTransformersNeed2024} encodes the input image into patch embeddings (\protect\tikz[baseline=0.0]\protect\draw[fill=dgray](0,0)rectangle(0.2,0.2);). The CLS token (\protect\tikz[baseline=0.0]\protect\draw[fill=dblue](0,0)rectangle(0.2,0.2);) and register tokens (\protect\tikz[baseline=0.0]\protect\draw[fill=dpurple](0,0)rectangle(0.2,0.2);) are discarded. The FlatDINO encoder---a ViT with learnable embedding tokens (\protect\tikz[baseline=0.0]\protect\draw[pattern={Lines[angle=45, distance=1.5pt, line width=0.6pt]}, pattern color=dgreen](0,0)rectangle(0.2,0.2);)---compresses the patch embeddings into a one-dimensional latent sequence (\protect\tikz[baseline=0.0]\protect\draw[fill=dgreen](0,0)rectangle(0.2,0.2);), achieving an 8$\times$ reduction in token count. The decoder, also a ViT with learnable embeddings (\protect\tikz[baseline=0.0]\protect\draw[pattern={Lines[angle=45, distance=1.5pt, line width=0.6pt]}, pattern color=dyellow](0,0)rectangle(0.2,0.2);), reconstructs the original DINOv2 patch embeddings (\protect\tikz[baseline=0.0]\protect\draw[fill=dyellow](0,0)rectangle(0.2,0.2);).}
  \label{fig:vae}
\end{figure*}

Diffusion models~\citep{sohl-dicksteinDeepUnsupervisedLearning2015, hoDenoisingDiffusionProbabilistic2020} have become the dominant approach for image generation, with latent diffusion~\citep{rombachHighResolutionImageSynthesis2022} reducing computational costs by operating in a compressed VAE latent space. Recent work has shown that diffusion models can also operate directly on self-supervised patch features rather than pixel-space latents. RAE~\citep{zhengDiffusionTransformersRepresentation2025} demonstrated that diffusing DINOv2~\citep{oquabDINOv2LearningRobust2024} patch embeddings yields faster convergence and competitive generation quality, with a learned decoder mapping features back to images (\cref{fig:vae}). This approach bypasses the reconstruction-generation trade-off inherent in pixel-trained VAEs~\citep{yaoReconstructionVsGeneration2025}, since DINOv2 features already encode semantically meaningful structure.

However, DINOv2 produces a dense two-dimensional grid of high-dimensional patch embeddings---256 patches of 768 dimensions at standard resolution. This representation is comparable in size to the original signal, offering no computational advantage over pixel-space diffusion. Because neighboring patches share substantial semantic content, this grid contains significant spatial redundancy.

We propose to compress this redundancy. Inspired by TiTok~\citep{yuImageWorth322024}, which showed that images can be encoded into just 32 one-dimensional tokens without sacrificing quality, we apply the same principle to DINOv2 features. Our method, FlatDINO, compresses 256 patch embeddings into 32 continuous tokens---an eightfold reduction in sequence length---while preserving reconstruction fidelity. Training a DiT with flow matching on this compact representation yields approximately $8\times$ faster inference.

Our contributions are as follows:
\begin{itemize}
    \item We introduce FlatDINO, which is the first method to compress self-supervised patch features into a one-dimensional latent representation, discarding the original two-dimensional spatial structure.
    \item We demonstrate that diffusion models trained on FlatDINO latents achieve competitive generation quality (gFID \textbf{1.85}) while reducing forward pass FLOPs by ${\sim}8\times$ compared to operating on full DINOv2 patches.
\end{itemize}

\section{Related Work}

\begin{figure*}[!t]
\centering
\includegraphics[width=0.95\textwidth]{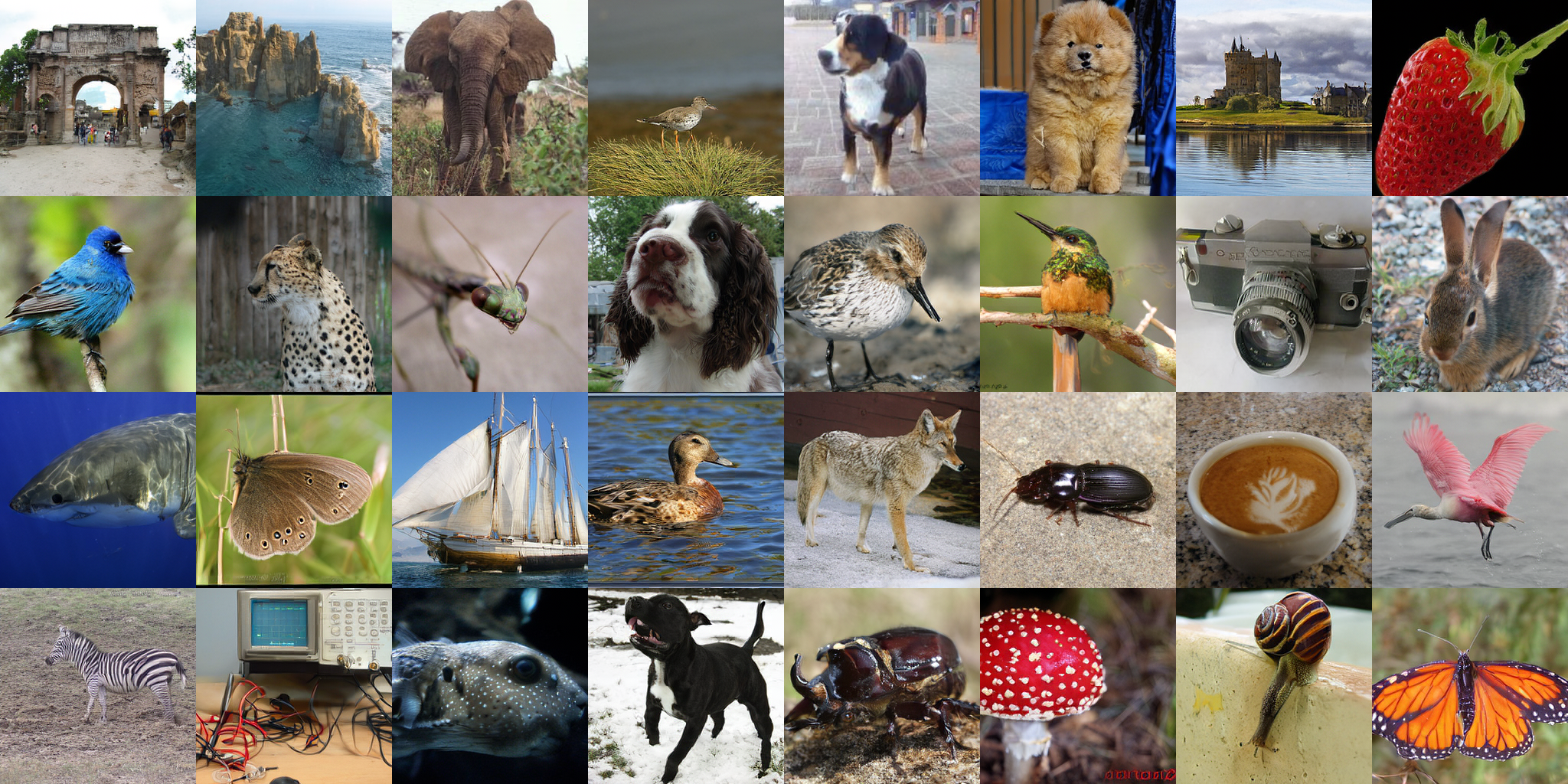}
\caption{Selected class-conditional samples from a DiT-XL model trained for 600 epochs on FlatDINO 32$\times$128 latents. Samples were generated using classifier-free guidance with an Euler sampler (50 steps). Despite the $8\times$ reduction in sequence length compared to RAE, FlatDINO produces diverse, high-fidelity images across a range of ImageNet classes.}
\label{fig:generated-samples}
\end{figure*}

\textbf{Latent diffusion and representation alignment.}
Diffusion models~\citep{songGenerativeModelingEstimating2020, songScoreBasedGenerativeModeling2021, hoDenoisingDiffusionProbabilistic2020, songDenoisingDiffusionImplicit2022} have emerged as the dominant paradigm for image generation. ADM~\citep{dhariwalDiffusionModelsBeat2021} first demonstrated that diffusion models can surpass GANs by introducing architectural improvements such as increased depth and attention modules at multiple image scales. However, pixel-space diffusion remains computationally demanding. Recent efforts to improve efficiency include SiD2~\citep{hoogeboomSimplerDiffusionSiD22025}, which adopts a U-ViT architecture---replacing some ResNet blocks with vision transformers---but still favors FLOP-heavy designs. PixelDiT~\citep{yuPixelDiTPixelDiffusion2025} introduces a dual-level architecture where transformer blocks first produce conditioning for $16 \times 16$ patches, followed by pixel transformer blocks that operate directly on pixel tokens. To manage the resulting sequence length, PixelDiT compacts $p^2$ tokens within each patch before global attention and expands them afterward; despite this compression, the model still processes large token sequences and requires substantial compute.

Latent diffusion models~\citep{rombachHighResolutionImageSynthesis2022} take a different approach, encoding images into a compressed VAE latent before applying diffusion. This enables high-resolution synthesis at reduced cost and has become the foundation for large-scale systems~\citep{podellSDXLImprovingLatent2023, esserScalingRectifiedFlow2024}. The Diffusion Transformer (DiT)~\citep{peeblesScalableDiffusionModels2023} replaced the U-Net backbone with a transformer, introducing adaptive layer normalization (adaLN-Zero) for conditioning injection and achieving state-of-the-art results.

Recent work has explored aligning diffusion with self-supervised representations. REPA~\citep{yuRepresentationAlignmentGeneration2024} adds a representation alignment loss that encourages the internal activations of the DiT to match SSL features such as those from DINOv2. The authors observed that DiT learns discriminative features in its intermediate layers during training, and that aligning these with already-discriminative SSL representations speeds up convergence considerably. \citet{singhWhatMattersRepresentation2025} later found that spatial structure is more important than global semantic information for this alignment; replacing REPA's MLP projection with a convolutional layer further improved convergence. REG~\citep{wuRepresentationEntanglementGeneration2025} showed that adding a token to the DiT input that learns to diffuse the DINOv2 CLS token, in combination with REPA, improves convergence speed further still. REPA-E~\citep{lengREPAEUnlockingVAE2025} introduced a method to finetune a pretrained VAE encoder with REPA's alignment loss, improving convergence over the original REPA, though REG remains faster.

RAE~\citep{zhengDiffusionTransformersRepresentation2025} takes a more direct approach by diffusing directly on DINOv2 patch features, bypassing the pixel-trained VAE entirely. Surprisingly, the authors demonstrated that a decoder can be trained to map DINOv2 patches back to images using the same recipe as modern pixel-reconstruction VAEs~\citep{rombachHighResolutionImageSynthesis2022}: a combination of pixel reconstruction loss, GAN loss~\citep{sauerStyleGANTUnlockingPower2023}, and perceptual loss~\citep{zhangUnreasonableEffectivenessDeep2018}. This achieves faster convergence and competitive generation quality, but operates on 256 high-dimensional patch tokens---the same sequence length as pixel-space latent diffusion. FlatDINO builds on RAE by compressing these features into a shorter sequence, reducing computational cost while preserving diffusability.

\textbf{Self-supervised visual representations.}
Self-supervised learning (SSL) has emerged as a powerful paradigm for learning visual representations without manual annotation. Discriminative SSL methods learn by solving pretext tasks that distinguish between data instances or groups, and can be broadly categorized into contrastive, clustering-based, and self-distillation approaches~\citep{giakoumoglouReviewDiscriminativeSelfsupervised2025}. Contrastive methods such as SimCLR~\citep{chenSimpleFrameworkContrastive2020} and MoCo~\citep{heMomentumContrastUnsupervised2020} learn by pulling positive pairs together while pushing negatives apart, but require large batch sizes or memory banks. Self-distillation methods circumvent this by using only positive pairs: BYOL~\citep{grillBootstrapYourOwn2020} introduced a student-teacher framework with an exponential moving average (EMA) teacher, while DINO~\citep{caronEmergingPropertiesSelfSupervised2021} extended this to Vision Transformers and showed that self-distillation with centering produces features with strong clustering properties and emergent segmentation capabilities. DINOv2~\citep{oquabDINOv2LearningRobust2024} scaled this approach with curated data, longer training, and architectural improvements, yielding general-purpose visual features that transfer effectively across tasks without fine-tuning. We build on DINOv2 features due to their strong semantic structure and discriminative properties.

\textbf{Image tokenization.}
Generative models benefit from compact latent representations that reduce sequence length while preserving visual fidelity. VQ-VAE~\citep{vandenoordNeuralDiscreteRepresentation2017} pioneered discrete image tokenization through vector quantization, and VQGAN~\citep{esserTamingTransformersHighResolution2021} improved quality with adversarial training, enabling masked generation methods like MaskGIT~\citep{changMaskGITMaskedGenerative2022} and MAGVIT~\citep{yuMAGVITMaskedGenerative2023}. Continuous tokenizers used in latent diffusion~\citep{rombachHighResolutionImageSynthesis2022} offer smoother latents suited to denoising objectives. These approaches share a common design choice: preserving the two-dimensional spatial layout of the image in the token grid.

TiTok~\citep{yuImageWorth322024} demonstrated that this is unnecessary, encoding images into a one-dimensional sequence of just 32 tokens without sacrificing quality. TA-TiTok~\citep{kimDemocratizingTexttoImageMasked2025} scaled this approach and introduced continuous latents for diffusion, while FlexTok~\citep{bachmannFlexTokResamplingImages2025} showed that 1D sequences can vary in length based on image complexity. FlatDINO extends this paradigm from pixels to features: we compress DINOv2's two-dimensional patch grid into a one-dimensional latent sequence, enabling efficient diffusion on self-supervised representations.

\section{Method}

\begin{figure}[t]
\centering
\includegraphics[width=0.95\columnwidth]{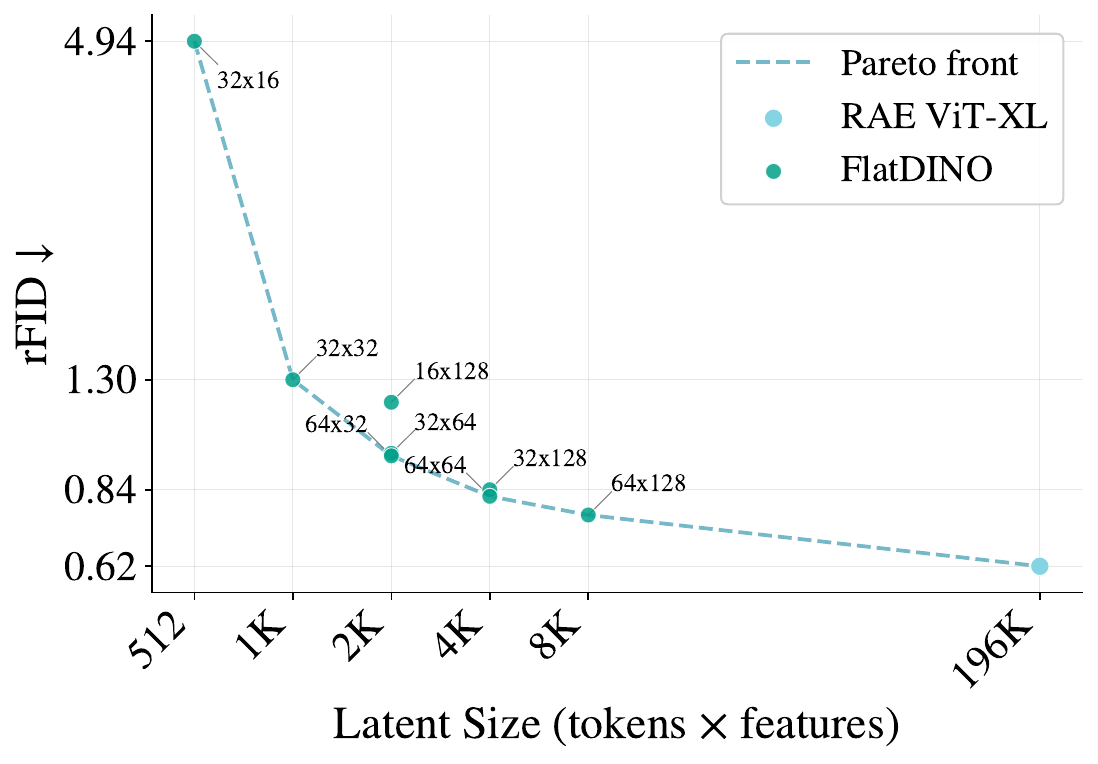}
\caption{Reconstruction quality (rFID, lower is better) versus total latent dimensionality for different token counts. All configurations were trained for 50 epochs; slightly better performance is expected with longer training. For a fixed latent size, configurations with more tokens consistently outperform those with fewer tokens but larger feature dimensions.}
\label{fig:latent-ablation}
\end{figure}

Autoencoders trained with pixel reconstruction losses tend to allocate latent capacity to low-level details such as textures and high-frequency content, which may not be optimal for generation~\citep{sanderLatents2025}. DINOv2 features, by contrast, already encode visual information according to semantic similarity~\citep{zhengDiffusionTransformersRepresentation2025}. By training an autoencoder to reconstruct this feature space, we optimize the latent to compress higher-level semantic content rather than pixel-level details.

Our method consists of three components: a variational autoencoder that compresses DINOv2 patch embeddings into a compact one-dimensional latent sequence, a pretrained decoder that maps DINOv2 features back to RGB images, and a flow-matching model that learns to generate novel latents.

\subsection{1D Autoencoder}

We compress the dense patch embeddings produced by DINOv2 with registers \citep{darcetVisionTransformersNeed2024} into a compact one-dimensional latent sequence using an autoencoder. We regularize the latent space with a small KL divergence penalty \citep{rombachHighResolutionImageSynthesis2022}: without such regularization, autoencoders can learn latents with arbitrary scale and distribution, making them incompatible with the diffusion process's noise schedule. This leads us to employ the variational autoencoder framework ($\beta$-VAE; \citet{kingmaAutoEncodingVariationalBayes2022, higginsBetaVAELearningBasic2017}).

Given an input image, DINOv2 produces $P$ patch embeddings (\protect\tikz[baseline=0.0]\protect\draw[fill=dgray](0,0)rectangle(0.2,0.2);) of dimension $D$. The encoder maps these embeddings to $T$ latent tokens (\protect\tikz[baseline=0.0]\protect\draw[fill=dgreen](0,0)rectangle(0.2,0.2);) of dimension $d$, discarding the original two-dimensional spatial structure. The decoder reconstructs the $P$ patch embeddings (\protect\tikz[baseline=0.0]\protect\draw[fill=dyellow](0,0)rectangle(0.2,0.2);) from the latent tokens, providing a consistent interface for downstream applications (\cref{fig:vae}). We refer to this model as FlatDINO. In our experiments we use DINOv2-B/14, which produces $P = 256$ patches of dimension $D = 768$.

Both encoder and decoder are implemented as Vision Transformers \citep{dosovitskiyImageWorth16x162021}, following the ViT-B and ViT-L architectures respectively. Inspired by \citet{kimDemocratizingTexttoImageMasked2025}, the encoder prepends $T$ learnable register tokens to the input patch sequence; after processing, only these registers are retained as the latent representation. The decoder similarly uses learnable registers as queries, which attend to the latent tokens and produce the reconstructed patch embeddings.

Training is performed on ImageNet-1k \citep{dengImageNetLargescaleHierarchical2009} using random cropping and horizontal flipping as data augmentation. We optimize with AdamW \citep{loshchilovDecoupledWeightDecay2019} using a batch size of 512 and a warmup-stable-decay (WSD) learning rate schedule \citep{huMiniCPMUnveilingPotential2024} with peak learning rate $10^{-4}$. In preliminary experiments, WSD yielded slightly better reconstruction quality than cosine decay for the same training budget, while also allowing seamless resumption for extended training. The model is trained using the standard VAE objective:
\begin{equation}
\mathcal{L} = \mathbb{E}_{q_\phi(z|x)}\left[-\log p_\theta(x | z)\right] + \beta \, D_{\mathrm{KL}}\left(q_\phi(z|x) \| p(z)\right),
\end{equation}
where $q_\phi$ and $p_\theta$ denote the encoder and decoder distributions, $x$ represents the DINOv2 patch embeddings, and $z \in \mathbb{R}^{T \times d}$ denotes the latent tokens. The KL weight $\beta$ is normalized by the latent dimensionality to ensure consistent regularization pressure across configurations; we set $\beta \propto 1/(T \cdot d)$ with a reference value of $\beta = 10^{-6}$ at $T \cdot d = 512$, following \citet{kimDemocratizingTexttoImageMasked2025}. We found that larger values of $\beta$ caused the autoencoder to collapse, producing latents that failed to preserve reconstruction fidelity. The $\beta$ values used for each latent dimensionality are listed in \cref{sec:kl-scaling}, along with the mathematical derivation.

Complete training details, including learning rate schedules and architectural hyperparameters, are provided in \cref{sec:flatdino-details}.

\subsection{Decoding to Images}

Since FlatDINO operates in DINOv2 feature space, both reconstructed and generated outputs are patch embeddings rather than RGB images. Visualizing these outputs requires inverting the DINOv2 representation---a non-trivial task, as self-supervised features prioritize semantic structure over low-level pixel details. Indeed, training a pixel decoder from SSL features was considered impractical due to invariances learned during pretraining (e.g., color jittering), but \citet{zhengDiffusionTransformersRepresentation2025} demonstrated that accurate reconstruction is possible for in-distribution images. We adopt their pretrained ViT-XL decoder, which remains frozen throughout our experiments. All image-space metrics (rFID, gFID) are computed on its outputs. We note that out-of-distribution images may exhibit color distortions (\cref{sec:ood-decoding}).

\subsection{Latent Generation}

To generate novel images, we train a generative model on the FlatDINO latent space using flow matching \citep{lipmanFlowMatchingGenerative2023}. Flow matching learns a velocity field $v_\theta(z_t, t)$ that transports samples from a simple prior distribution $p_0$ (in our case a standard Gaussian) to the data distribution $p_1$ along straight paths. Given a data sample $z_1 \sim p_1$ and noise $z_0 \sim \mathcal{N}(0, I)$, the interpolant at time $t \in [0, 1]$ is defined as:
\begin{equation}
z_t = (1 - t) z_0 + t z_1.
\end{equation}
The velocity field is trained to match the conditional flow:
\begin{equation}
\mathcal{L}_{\mathrm{FM}} = \mathbb{E}_{t, z_0, z_1} \left\| v_\theta(z_t, t) - (z_1 - z_0) \right\|^2.
\end{equation}
At inference, samples are generated by integrating the learned velocity field from $t = 0$ to $t = 1$ using an ODE solver, transforming Gaussian noise into latent tokens that can be decoded to DINOv2 patch embeddings.

We follow the training protocol described in RAE \citep{zhengDiffusionTransformersRepresentation2025}, parameterizing the velocity field with LightningDiT \citep{yaoReconstructionVsGeneration2025}, an efficient variant of DiT \citep{peeblesScalableDiffusionModels2023}. The model operates directly on the one-dimensional FlatDINO latent sequence using learned positional embeddings. We train with a batch size of 1024 using AdamW ($\beta_1 = 0.9$, $\beta_2 = 0.95$) with a constant learning rate of $2 \times 10^{-4}$, and apply exponential moving average (EMA) with decay 0.9995. Operating on the compressed 32-token FlatDINO representation rather than the full 256 DINOv2 patches reduces the sequence length by $8\times$, yielding substantial computational savings during both training and inference (\cref{sec:flops}). Generation quality is evaluated using 50 Euler integration steps, unless specified otherwise.

\section{Experiments}

\begin{table*}[t]
\centering
\caption{Comparison of image generation methods on ImageNet 256$\times$256.}
\label{tab:generation-comparison}
\begin{footnotesize}
\begin{tabular}{lcccccccc}
\toprule
& & & \multicolumn{3}{c}{\textbf{w/o CFG}} & \multicolumn{3}{c}{\textbf{w/ CFG}} \\
\cmidrule(lr){4-6} \cmidrule(lr){7-9}
\textbf{Method} & \textbf{Epochs} & \textbf{Params} & \textbf{gFID $\downarrow$} & \textbf{Prec. $\uparrow$} & \textbf{Rec. $\uparrow$} & \textbf{gFID $\downarrow$} & \textbf{Prec. $\uparrow$} & \textbf{Rec. $\uparrow$} \\
\midrule
\multicolumn{9}{c}{\textit{\textbf{Autoregressive}}} \\
\midrule
VAR \citep{tianVisualAutoregressiveModeling2024} & 350 & 2.0B & 1.92 & 0.82 & 0.59 & 1.73 & 0.82 & 0.60 \\
MAR \citep{liAutoregressiveImageGeneration2024} & 800 & 943M & 2.35 & 0.79 & 0.62 & 1.55 & 0.81 & 0.62 \\
xAR \citep{renNextTokenNextXPrediction2025} & 800 & 1.1B & -- & -- & -- & 1.24 & 0.83 & 0.64 \\
\midrule
\multicolumn{9}{c}{\textit{\textbf{Pixel Diffusion}}} \\
\midrule
ADM \citep{dhariwalDiffusionModelsBeat2021} & 400 & 554M & 10.94 & 0.69 & 0.63 & 3.94 & 0.83 & 0.53 \\
PixelFlow \citep{chenPixelFlowPixelSpaceGenerative2025} & 320 & 677M & -- & -- & -- & 1.98 & 0.81 & 0.60 \\
PixNerd \citep{wangPixNerdPixelNeural2025} & 160 & 700M & -- & -- & -- & 2.15 & 0.79 & 0.59 \\
\midrule
\multicolumn{9}{c}{\textit{\textbf{Latent Diffusion with VAE}}} \\
\midrule
DiT \citep{peeblesScalableDiffusionModels2023} & 1400 & 675M & 9.62 & 0.67 & 0.67 & 2.27 & 0.83 & 0.57 \\
MaskDiT \citep{zhengFastTrainingDiffusion2024} & 1600 & 675M & 5.69 & 0.74 & 0.60 & 2.28 & 0.80 & 0.61 \\
SiT \citep{maSiTExploringFlow2024} & 1400 & 675M & 8.61 & 0.68 & 0.67 & 2.06 & 0.82 & 0.59 \\
MDTv2 \citep{gaoMDTv2MaskedDiffusion2024} & 1080 & 675M & -- & -- & -- & 1.58 & 0.79 & 0.65 \\
\cmidrule{1-9}
\multirow{2}{*}{VA-VAE \citep{yaoReconstructionVsGeneration2025}} & 80 & 675M & 4.29 & -- & -- & -- & -- & -- \\
 & 800 & 675M & 2.17 & 0.77 & 0.65 & 1.35 & 0.79 & 0.65 \\
\cmidrule{1-9}
\multirow{2}{*}{REPA \citep{yuRepresentationAlignmentGeneration2025}} & 80 & 675M & 7.90 & 0.70 & 0.65 & -- & -- & -- \\
 & 800 & 675M & 5.78 & 0.70 & 0.68 & 1.29 & 0.79 & 0.64 \\
\cmidrule{1-9}
\multirow{2}{*}{REPA-E \citep{lengREPAEUnlockingVAE2025}} & 80 & 675M & 3.46 & 0.77 & 0.63 & 1.67 & 0.80 & 0.63 \\
 & 800 & 675M & 1.70 & 0.77 & 0.66 & 1.15 & 0.79 & 0.66 \\
\cmidrule{1-9}
\multirow{2}{*}{DDT \citep{wangDDTDecoupledDiffusion2025}} & 80 & 675M & 6.62 & 0.69 & 0.67 & 1.52 & 0.78 & 0.63 \\
 & 400 & 675M & 6.27 & 0.68 & 0.69 & 1.26 & 0.79 & 0.65 \\
\midrule
\multicolumn{9}{c}{\textit{\textbf{RAE}} \citep{zhengDiffusionTransformersRepresentation2025}} \\
\midrule
DiT-XL (DINOv2-S) & 800 & 676M & 1.87 & 0.80 & 0.63 & 1.41 & 0.80 & 0.63 \\
$\text{DiT}^{\text{DH}}$-XL (DINOv2-B) & 800 & 839M & 1.51 & 0.79 & 0.63 & 1.13 & 0.78 & 0.67 \\
\midrule
\multicolumn{9}{c}{\textit{\textbf{FlatDINO}}} \\
\midrule
DiT-XL (32$\times$128) & 600 & 676M  & 3.34 & 0.77 & 0.61 & 1.85 & 0.78 & 0.63 \\
DiT-XL (32$\times$128) & 800 & 676M  & 3.21 & 0.77 & 0.62 & 1.80 & 0.78 & 0.62 \\
\bottomrule
\end{tabular}
\end{footnotesize}
\end{table*}

\subsection{Latent Shape Selection}

We ablate the latent representation by varying the number of tokens (16, 32, 64) and the per-token feature dimension (16, 32, 64, 128), training each configuration for 50 epochs on ImageNet-1k. Reconstruction quality is measured using rFID computed against DINOv2 patch features on the validation set. To obtain RGB images for visualization, we pass latent tokens sequentially through the FlatDINO decoder (recovering DINOv2 patch embeddings) and then through the RAE decoder (mapping patches to pixels).

As shown in \cref{fig:latent-ablation}, for a given total latent dimensionality, allocating capacity to more tokens rather than larger per-token features yields better reconstruction. At 2048 total dimensions, for instance, the 64-token configuration (64$\times$32) achieves an rFID of 0.96, compared to 0.97 for 32 tokens (32$\times$64) and 1.19 for 16 tokens (16$\times$128). The gap between 64 and 32 tokens is marginal, but reducing to 16 tokens incurs a notable quality drop. This pattern suggests that spatial coverage provided by additional tokens is more valuable than richer per-token representations, though returns diminish beyond 32 tokens. The sharp quality degradation when reducing to 16 tokens is particularly striking---we investigate what drives this transition in \cref{sec:token-ablation}.

Based on these results, we select 32 tokens as the default configuration, offering a favorable trade-off between reconstruction fidelity and sequence length. We train the 32$\times$64 and 32$\times$128 configurations for 150 epochs (\cref{tab:latent-ablation}).

\begin{table}[t]
\centering
\caption{Reconstruction quality (rFID, computed on RAE-decoded images \citep{zhengDiffusionTransformersRepresentation2025}) for selected FlatDINO configurations trained for 150 epochs. Compression ratio is relative to DINOv2's 256$\times$768 representation.}
\label{tab:latent-ablation}
\begin{small}
\begin{tabular}{lcccc}
\toprule
\textbf{Model} & \textbf{Tokens} & \textbf{Dim} & \textbf{Compression} & \textbf{rFID} \\
\midrule
RAE (baseline) & 256 & 768 & 1$\times$ & 0.62 \\
\midrule
FlatDINO 32$\times$128 & 32 & 128 & 48$\times$ & 0.77 \\
FlatDINO 32$\times$64 & 32 & 64 & 96$\times$ & 0.87 \\
\bottomrule
\end{tabular}
\end{small}
\end{table}

\subsection{Token Ablation}
\label{sec:token-ablation}

What changes in the learned representation when we reduce from 32 to 16 tokens? To answer this, we analyze what visual information each token encodes by zeroing out individual tokens and measuring the resulting change in reconstruction. For each token, we compute the L2 distance between the original and ablated reconstructions in DINOv2 patch embedding space, averaged over 10,000 ImageNet validation images. This procedure yields a spatial heatmap indicating which image regions are most affected by each token.

\begin{figure}[t]
\centering
\includegraphics[width=0.95\columnwidth]{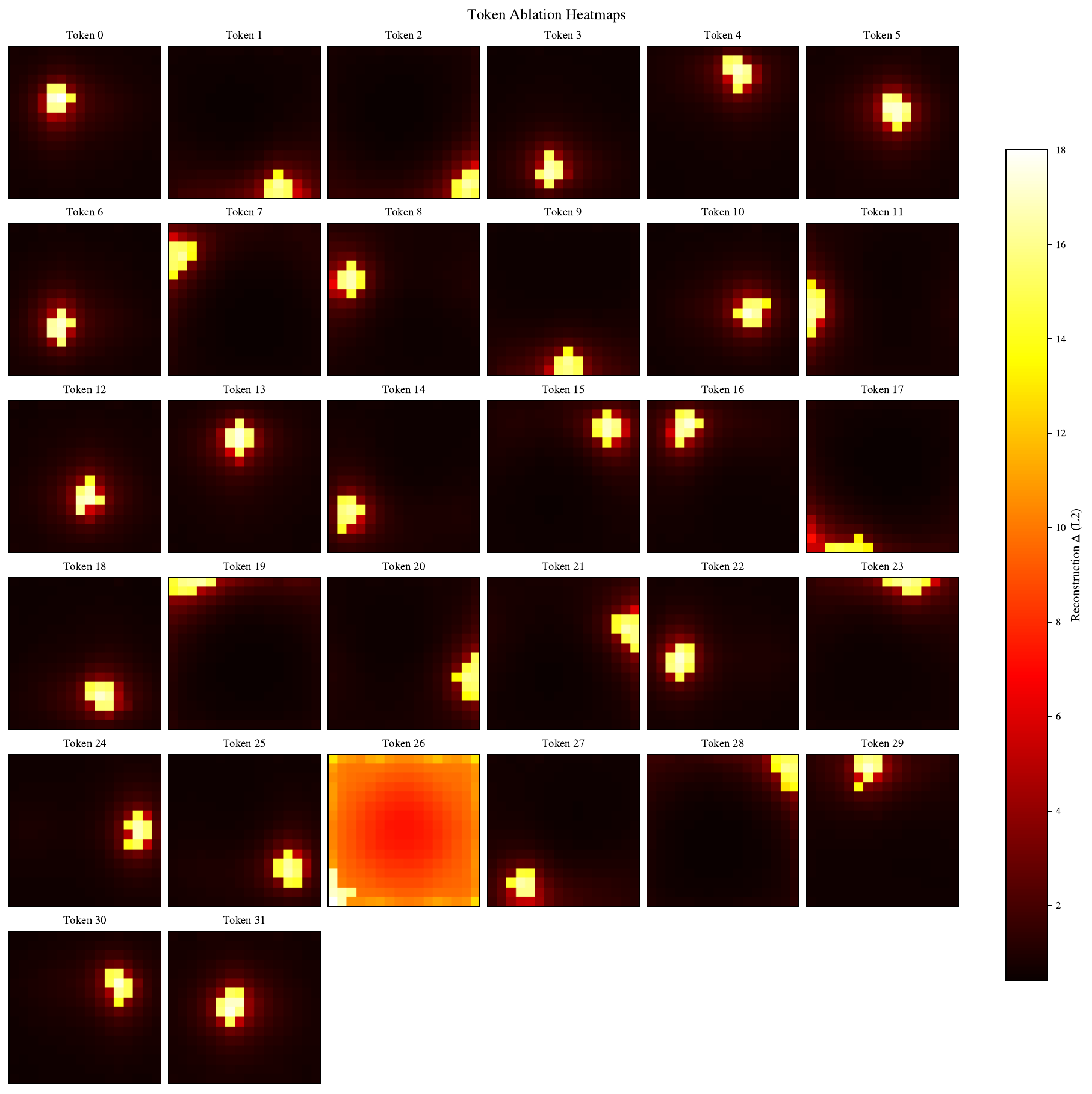}
\caption{Token ablation heatmaps for the 32-token FlatDINO model (32$\times$64). Each subplot shows the mean reconstruction error when that token is zeroed out, averaged over 10,000 ImageNet validation images. Most tokens learn localized blob-like receptive fields, suggesting spatial partitioning. Token 26 is an exception, showing diffuse image-wide influence consistent with encoding background regions.}
\label{fig:token-ablation}
\end{figure}

\begin{figure*}[!htbp]
\centering
\setlength{\tabcolsep}{1pt}
\begin{tabular}{c@{\hspace{4pt}}cccccc}
  & Original & Token 00 & Token 05 & Token 06 & Token 24 & Token 26 \\[2pt]
  & \ablationoriginal{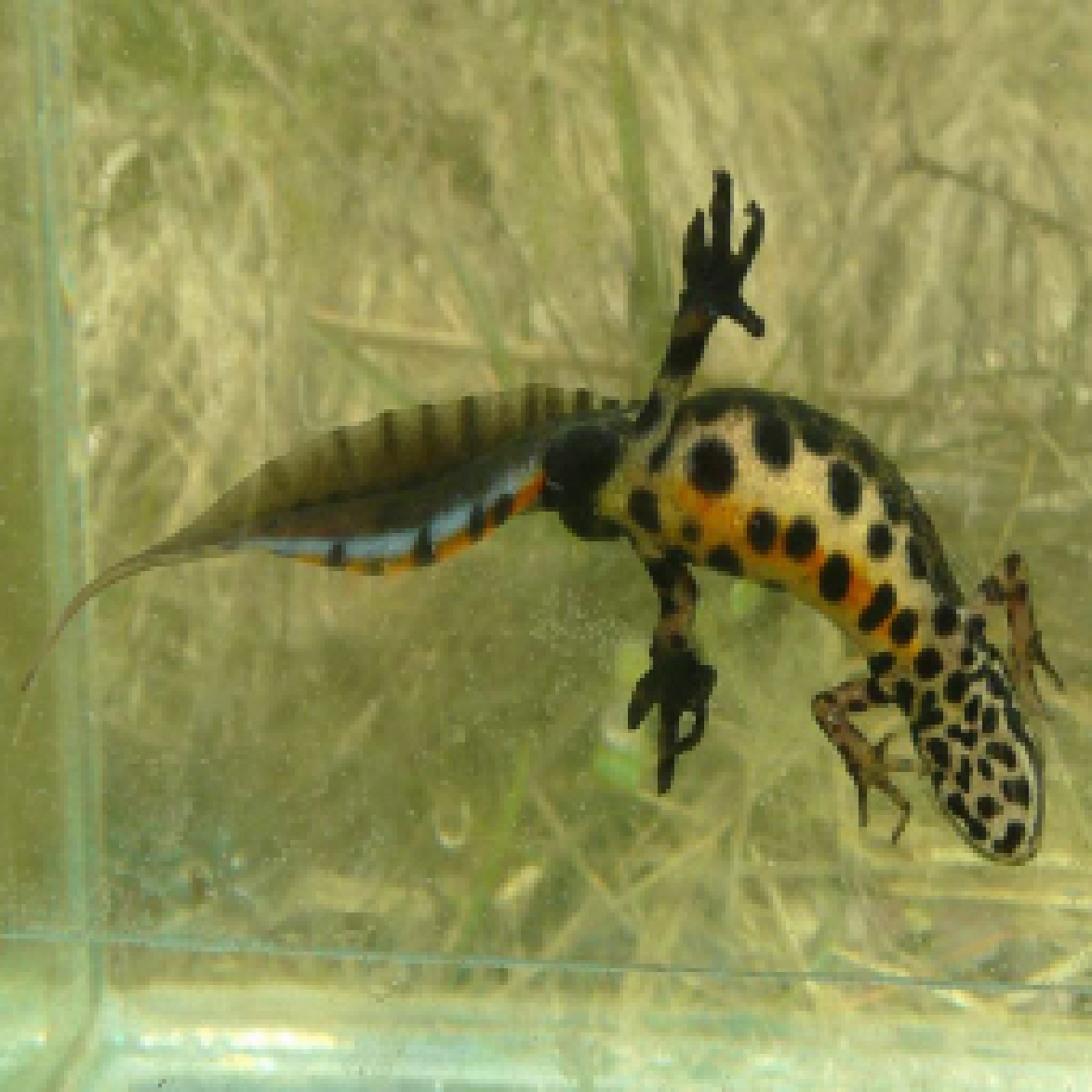} & \ablationtoken{00}{00} & \ablationtoken{00}{05} & \ablationtoken{00}{06} & \ablationtoken{00}{24} & \ablationtoken{00}{26} \\[1pt]
  & \ablationoriginal{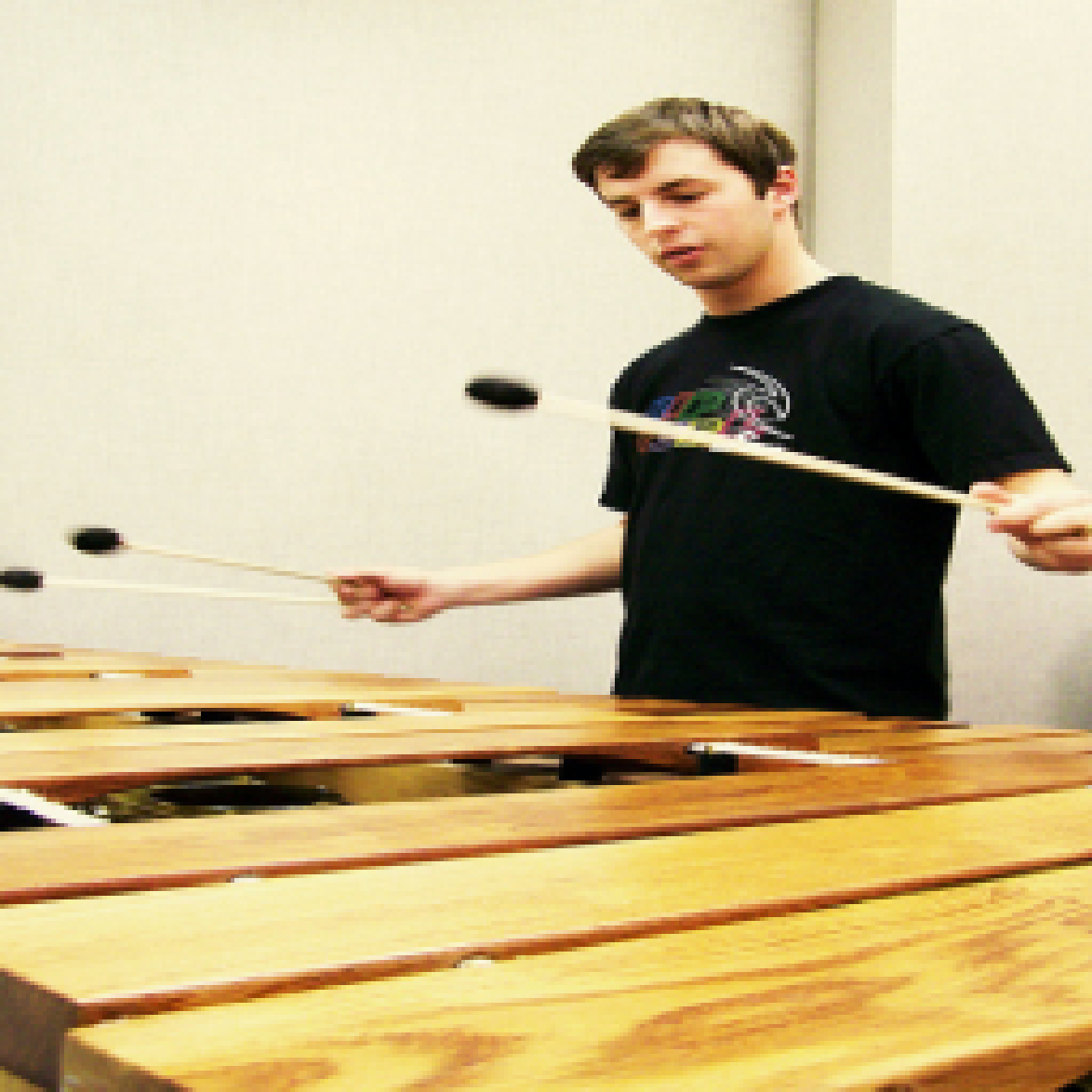} & \ablationtoken{03}{00} & \ablationtoken{03}{05} & \ablationtoken{03}{06} & \ablationtoken{03}{24} & \ablationtoken{03}{26} \\[1pt]
  & \ablationoriginal{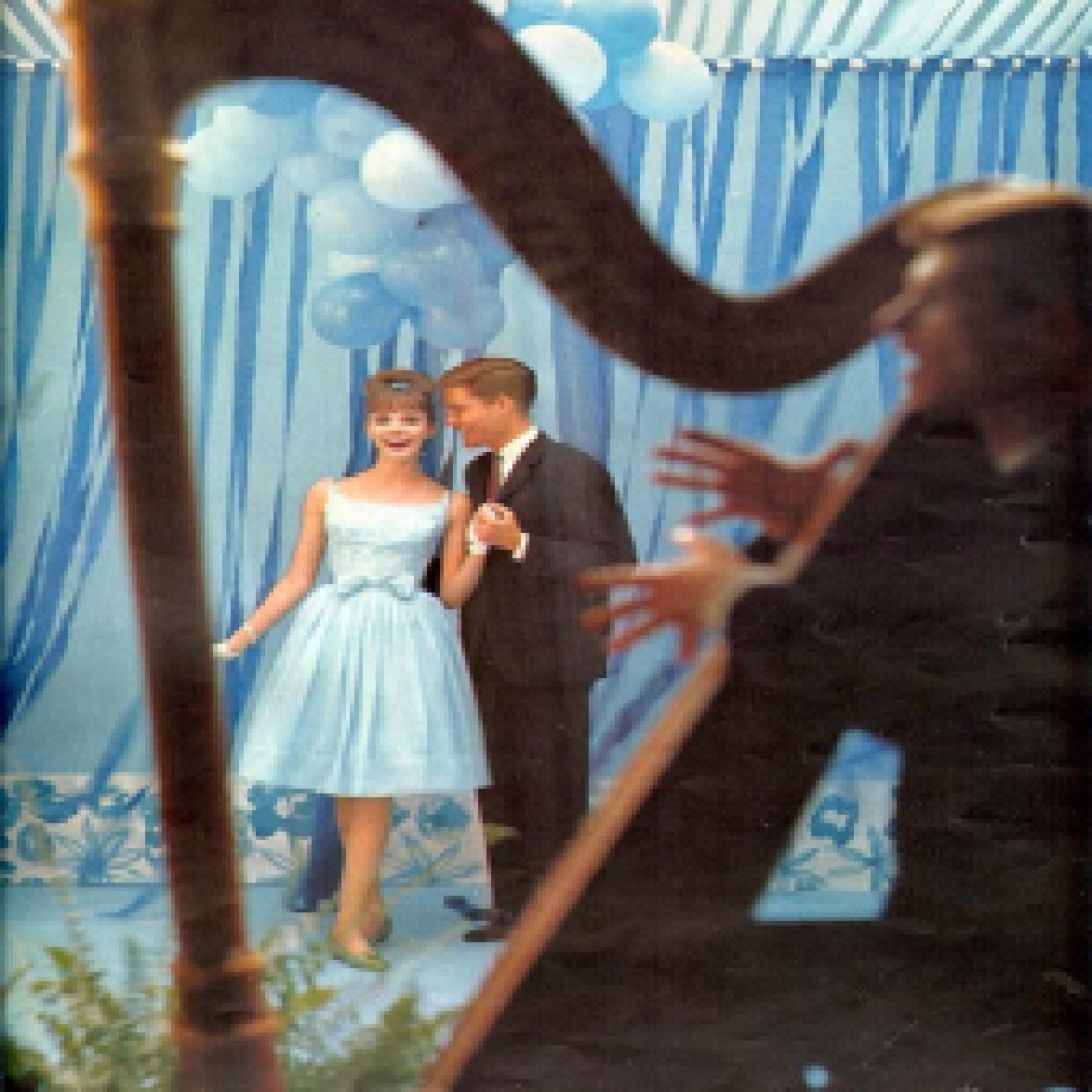} & \ablationtoken{04}{00} & \ablationtoken{04}{05} & \ablationtoken{04}{06} & \ablationtoken{04}{24} & \ablationtoken{04}{26} \\[1pt]
  & \ablationoriginal{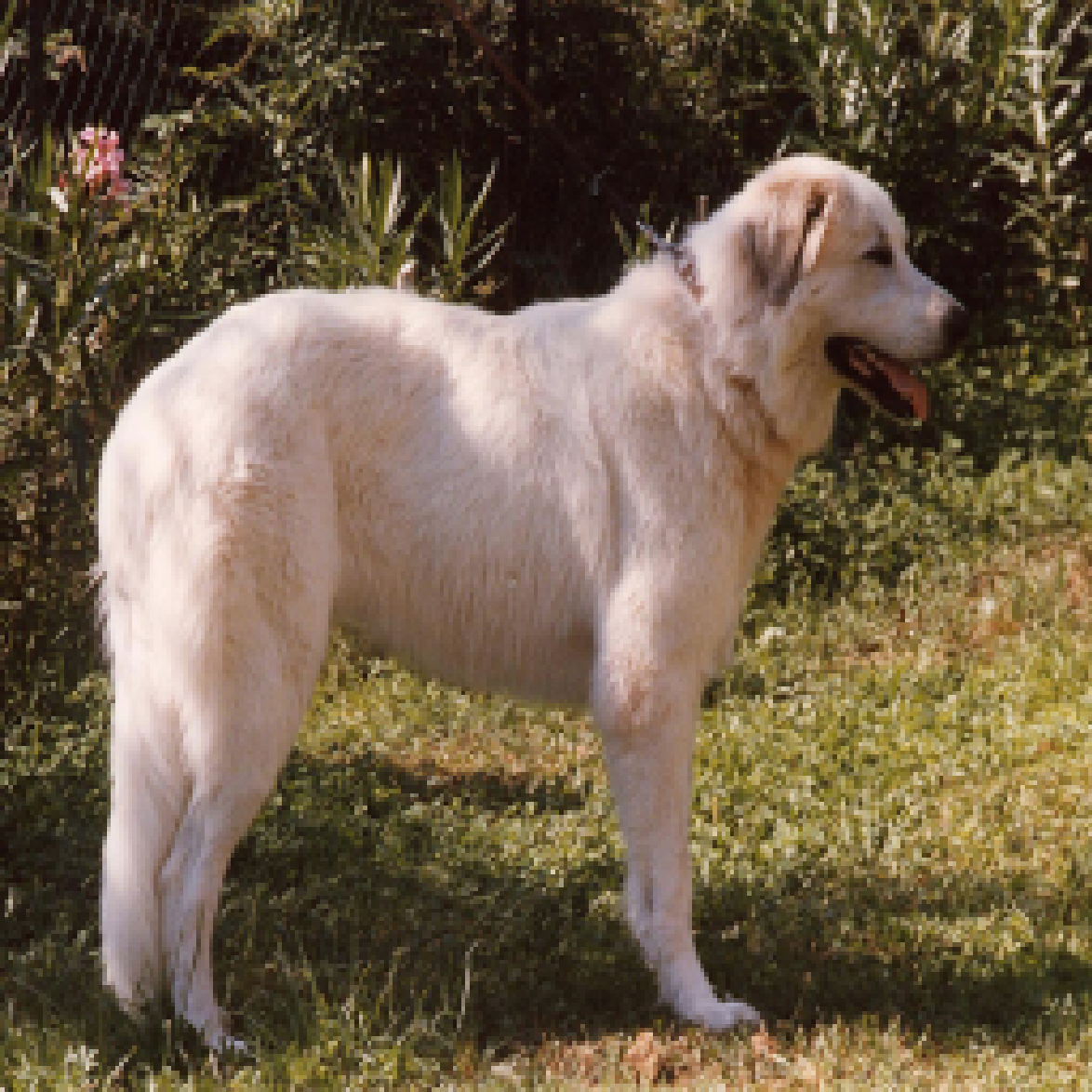} & \ablationtoken{06}{00} & \ablationtoken{06}{05} & \ablationtoken{06}{06} & \ablationtoken{06}{24} & \ablationtoken{06}{26} \\
\end{tabular}
\caption{Qualitative token ablation examples showing RAE-decoded reconstructions when individual tokens are zeroed out. Tokens 00, 05, 06, and 24 produce localized artifacts, while token 26 affects background regions across the image.}
\label{fig:token-ablation-images}
\end{figure*}

\Cref{fig:token-ablation} shows the spatial organization of the 32-token FlatDINO model. Most tokens exhibit localized receptive fields, each affecting a contiguous ``blob'' of patches in the spatial grid. This suggests that FlatDINO learns a form of spatial partitioning, with different tokens specializing in different image regions. One notable exception is token 26, which produces a diffuse, image-wide response. Visualizing the effect of ablating this token on individual images (\cref{fig:token-ablation-images}) reveals that its influence concentrates on background regions rather than foreground objects, suggesting it specializes in encoding scene context (see \cref{sec:token-ablation-configs} for further analysis).

Why does FlatDINO learn localized blobs rather than more semantic groupings? One possible explanation is that compression along the feature dimension is difficult, while spatial compression is easier. To test this, we measure the linear compressibility of DINOv2 features via PCA (\cref{tab:intrinsic_dim}). While pixel patches require only 21 of 588 dimensions to explain 95\% of variance---a 28$\times$ compression---DINO features require 594 of 768 dimensions, permitting merely 1.3$\times$ compression. This suggests that DINO has already decorrelated its outputs; little linear redundancy remains along the feature dimension.

\begin{table}[t]
\centering
\caption{Linear compressibility of DINO features versus pixel patches, measured via PCA on 2.57M patches from ImageNet validation images.}
\label{tab:intrinsic_dim}
\begin{small}
\begin{tabular}{lcc}
\toprule
\textbf{Metric} & \textbf{DINO} & \textbf{Pixels} \\
\midrule
Feature dimension & 768 & 588 \\
Dims for 95\% variance & 594 & 21 \\
Compression ratio @ 95\% & 1.3$\times$ & 28.0$\times$ \\
\bottomrule
\end{tabular}
\end{small}
\end{table}

\begin{figure}[t]
\centering
\includegraphics[width=0.95\columnwidth]{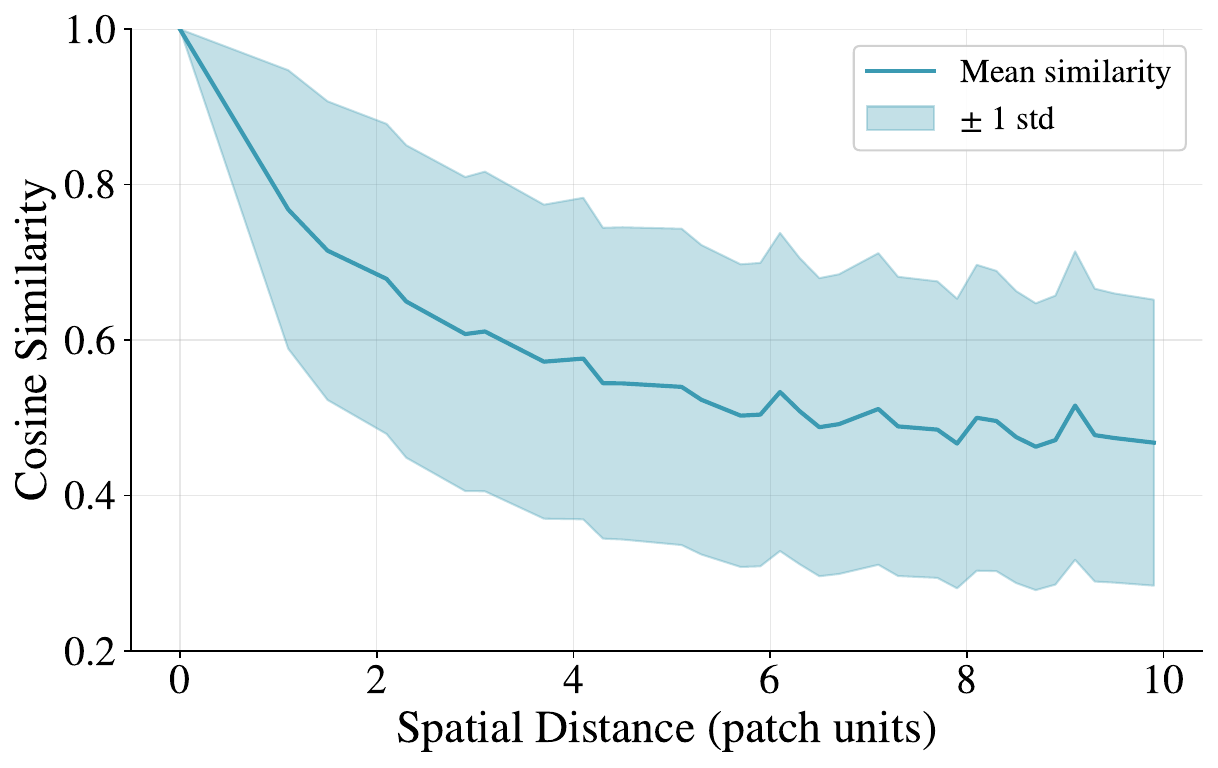}
\caption{Cosine similarity between DINOv2-B patch embeddings as a function of spatial distance (in patch units), averaged over ImageNet validation images. Nearby patches share more information than distant ones, which may explain why FlatDINO learns spatially localized receptive fields.}
\label{fig:spatial-redundancy}
\end{figure}

However, spatial redundancy remains. We compute the average cosine similarity between patch embeddings as a function of their spatial distance (\cref{fig:spatial-redundancy}). Similarity decreases monotonically with distance, confirming that nearby patches share substantially more information than distant ones. This suggests that localized receptive fields may be an efficient encoding strategy: by grouping spatially adjacent patches, each token can exploit their shared information for compression.

Interestingly, the 16-token model exhibits qualitatively different behavior (\cref{sec:token-ablation-configs}): rather than encoding localized blobs, most tokens learn to represent entire horizontal stripes of the image. This suggests a phase transition in the learned representation as the number of tokens decreases. We hypothesize that this transition explains the sharp degradation in reconstruction quality observed when reducing from 32 to 16 tokens (\cref{fig:latent-ablation}): horizontal stripes may be a less efficient encoding strategy than localized blobs, as they cannot exploit the two-dimensional spatial correlations present in natural images. With sufficient tokens, the model can tile the image with compact, localized receptive fields; with too few tokens, it reverts to a coarser one-dimensional partitioning that sacrifices spatial precision.

\subsection{Latent Diffusion}

We initially attempted to train a generative model on 32$\times$768 latents, preserving the full DINOv2 feature dimension. However, we found that diffusion models were unable to converge on this high-dimensional latent space, motivating us to compress the per-token feature dimension alongside the token count. We then conducted preliminary experiments with both 32$\times$64 and 32$\times$128 configurations, observing that 32$\times$128 converged faster. Examining the token ablation of the 32$\times$64 model (\cref{fig:token-ablation-32x64}), we find that five tokens learn to encode global rather than local information, compared to only one in 32$\times$128. We hypothesize that these additional global tokens deteriorate the structure of the latent space for diffusion, but leave a full investigation for future work.

We train a DiT-XL model on FlatDINO 32$\times$128 latents for 600 epochs, following the training recipe of LightningDiT \citep{yaoReconstructionVsGeneration2025}. During both training and inference, we apply a time shifting transformation $t' = t / (\kappa - (\kappa - 1) t)$ with $\kappa = 3$, which biases the diffusion process toward later timesteps where fine details emerge. RAE employs a dimensionality-dependent shift derived from their latent statistics; however, we found through preliminary experiments that this formulation does not transfer well to the FlatDINO latent space, necessitating our fixed $\kappa$ value.

\Cref{tab:generation-comparison} compares FlatDINO to prior work on ImageNet 256$\times$256 generation. FlatDINO achieves a gFID of 3.34 without classifier-free guidance (1.85 with CFG), demonstrating that the compressed latent space remains amenable to diffusion-based generation. For CFG, we apply guidance with weight 4.5 only during $t \in [0.225, 1.0]$, following the limited-interval approach of \citet{kynkaanniemiApplyingGuidanceLimited2024}. We use the same time shifting at inference as during training, and select these hyperparameters via a sweep over CFG weights and starting intervals (\cref{sec:cfg-sweep}). \Cref{fig:generated-samples} shows selected samples generated with classifier-free guidance. We note that the model has not fully converged and would benefit from additional training; however, due to computational constraints, we leave longer training runs for future work. We also note that our 600-epoch training is shorter than RAE's 800 epochs, which may account for part of the generation quality gap.

\section{Discussion and Future Work}

We have introduced FlatDINO, a variational autoencoder that compresses DINOv2 patch embeddings into a one-dimensional sequence of continuous tokens. By exploiting the spatial redundancy inherent in two-dimensional patch grids, FlatDINO achieves an eightfold reduction in token count while preserving reconstruction fidelity and retaining the semantic structure that makes self-supervised features amenable to diffusion-based generation.

Training a flow matching model on the FlatDINO latent space yields substantial savings in both compute and memory. The eightfold reduction in sequence length directly translates to faster training iterations and more efficient sampling, making diffusion on semantic representations more practical for resource-constrained settings without sacrificing the convergence benefits of operating in a semantically organized latent space.

We acknowledge that the generation quality of our current model, while competitive, does not yet match the state of the art achieved by methods operating on uncompressed DINOv2 features or by the latest autoregressive approaches. We attribute this gap partly to insufficient training duration---our model has not fully converged---and partly to the need for diffusion recipes specifically tailored to compressed semantic latents. In future work, we plan to investigate strategies for closing this gap, including extended training schedules, recent advances in diffusion architectures and sampling techniques, and the development of SSL autoencoders optimized jointly for reconstruction and generation. We believe that the efficiency-quality tradeoff demonstrated by FlatDINO represents a promising direction for scalable, semantically grounded image synthesis.

\section*{Impact Statement}

This paper presents work whose goal is to advance the field of Machine
Learning. There are many potential societal consequences of our work, none
which we feel must be specifically highlighted here.

\FloatBarrier

\bibliography{main,manual}
\bibliographystyle{icml2026}

\newpage
\appendix
\onecolumn

%
%

\section{Out-of-Distribution Decoding}
\label{sec:ood-decoding}

The RAE decoder \citep{zhengDiffusionTransformersRepresentation2025} was trained on ImageNet to invert DINOv2 representations back to RGB images. While it recovers colors accurately for in-distribution images, we observe that out-of-distribution inputs exhibit noticeable color distortions.

\begin{figure}[h]
\centering
\includegraphics[width=0.7\textwidth, trim=0 280 0 0, clip]{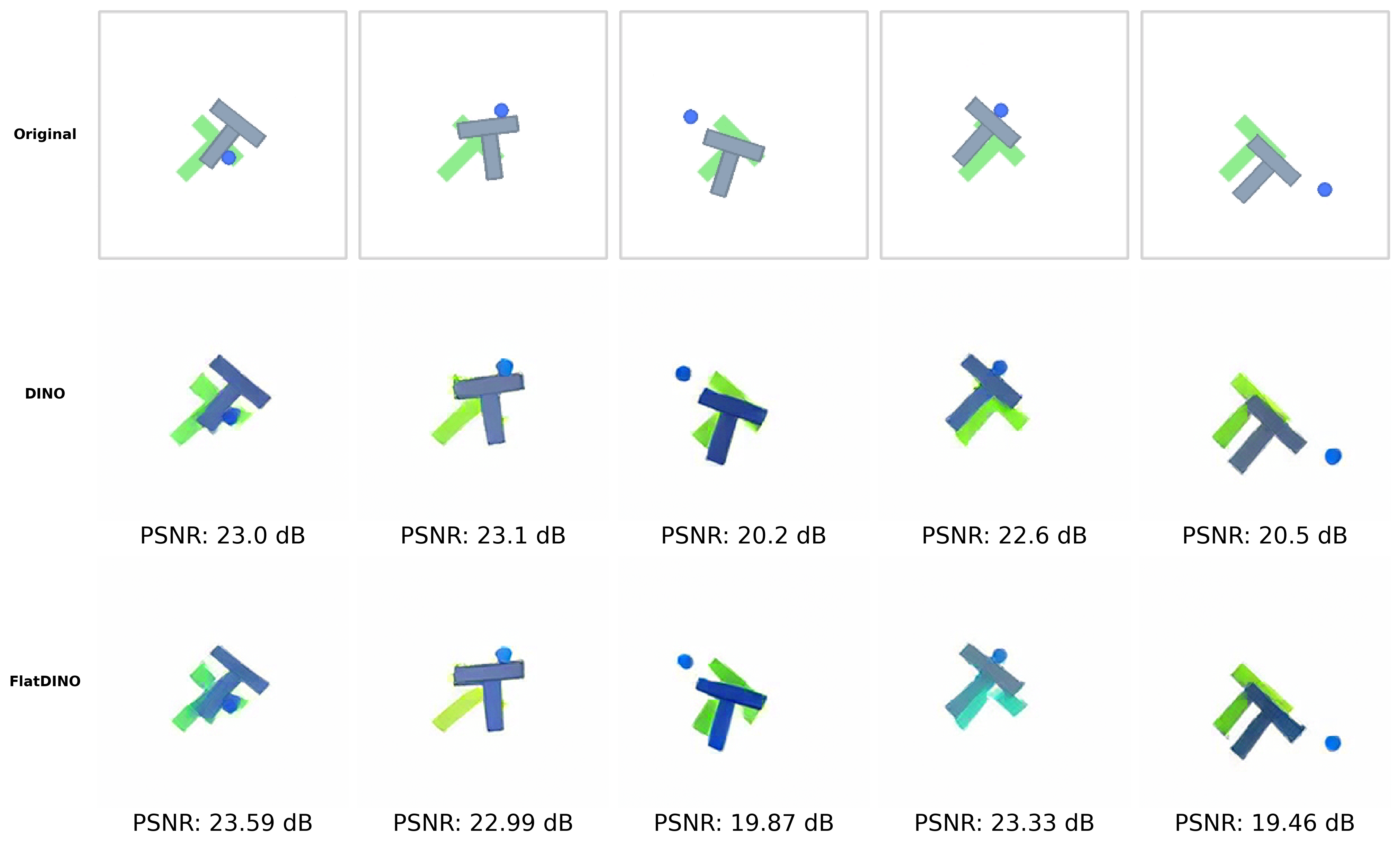}
\caption{DINOv2 encoding and RAE decoding of PushT frames. Top row: original frames. Bottom row: reconstructions. While spatial structure is preserved, colors are distorted due to the domain gap between PushT and the ImageNet-trained decoder.}
\label{fig:pusht-reconstruction}
\end{figure}

\Cref{fig:pusht-reconstruction} shows frames from the PushT task \citep{chiDiffusionPolicyVisuomotor2024}, a robotic manipulation benchmark where an agent must push a T-shaped block to a target pose. When these frames are encoded with DINOv2 and reconstructed with the RAE decoder, the spatial structure is preserved but the colors are inaccurate---the agent and block appear with shifted hues compared to the originals. We attribute this to the domain gap between PushT's synthetic rendering and ImageNet's natural images. This limitation does not affect our main experiments, which operate entirely within the ImageNet domain.

\FloatBarrier
\section{Latent Robustness to Noise}
\label{sec:noise-robustness}

To characterize how the compressed FlatDINO representations behave under perturbation, we evaluate reconstruction quality when Gaussian noise is added to the latent codes. This experiment probes the sensitivity of the decoder to deviations from the learned latent manifold---a property relevant both for understanding the geometry of the latent space and for assessing robustness during generation, where the diffusion process must traverse noisy intermediate states.

\begin{figure}[h]
\centering
\begin{minipage}{0.48\textwidth}
\centering
\includegraphics[width=\textwidth]{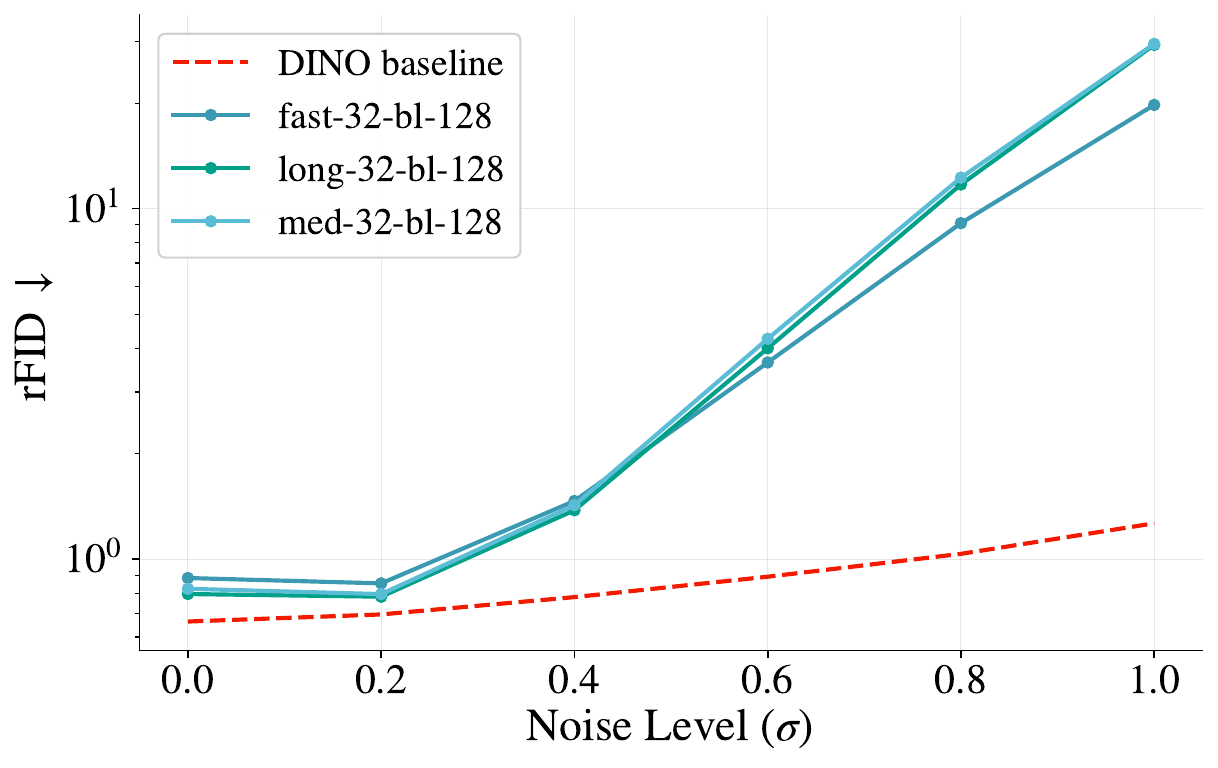}
\captionof{figure}{Reconstruction quality under latent perturbation for FlatDINO 32$\times$128. Reconstructions remain stable for $\sigma < 0.2$.}
\label{fig:noise-levels-128}
\end{minipage}
\hfill
\begin{minipage}{0.48\textwidth}
\centering
\includegraphics[width=\textwidth]{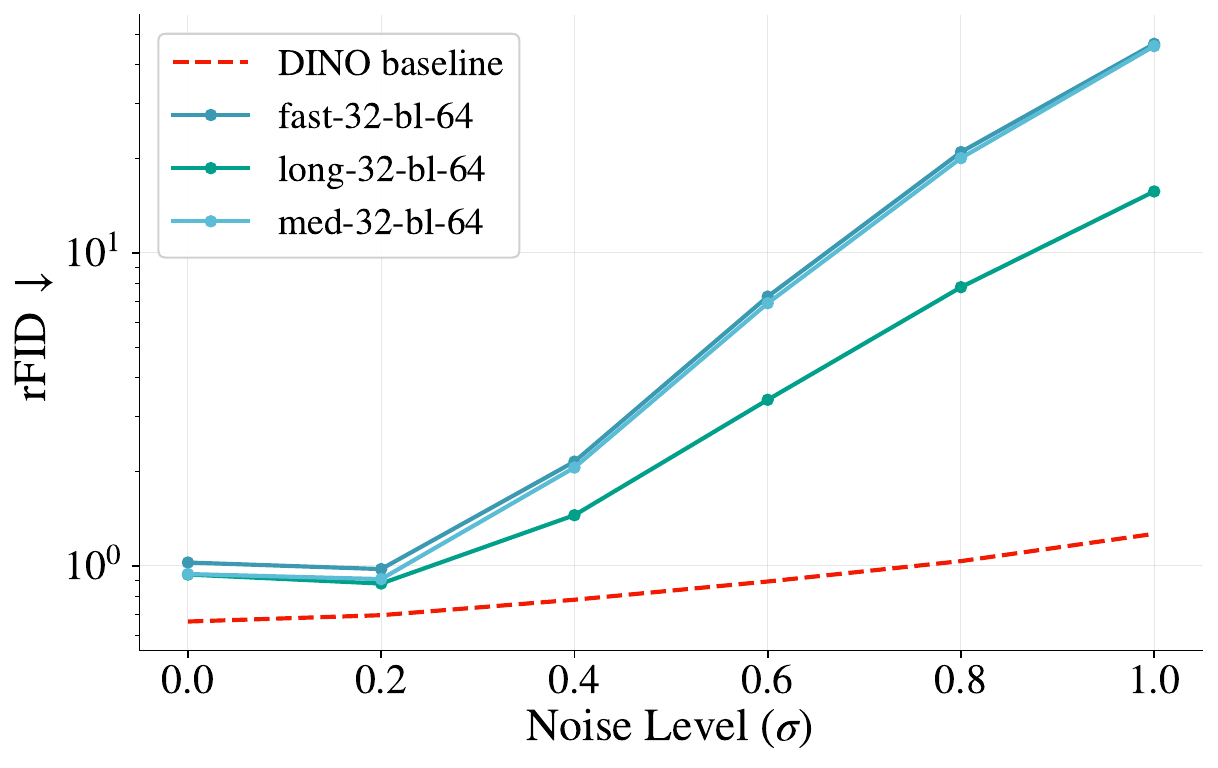}
\captionof{figure}{Reconstruction quality under latent perturbation for FlatDINO 32$\times$64. Similar robustness, with degradation beginning around $\sigma = 0.2$.}
\label{fig:noise-levels-64}
\end{minipage}
\end{figure}

We inject noise of varying magnitudes $\sigma \in [0, 1]$ into the latent representations and decode the perturbed codes back to images. \Cref{fig:noise-levels-128,fig:noise-levels-64} show the results for the 32$\times$128 and 32$\times$64 configurations, respectively. Both latent spaces exhibit graceful degradation: at low noise levels ($\sigma < 0.2$), reconstructions remain visually indistinguishable from the noise-free baseline, indicating that the decoder tolerates small perturbations without introducing artifacts.

However, at higher noise levels, reconstruction quality degrades more rapidly than when equivalent noise is applied directly to DINOv2 patch features before decoding. This accelerated degradation is expected: FlatDINO's compression concentrates information into fewer dimensions, so perturbations in the compressed space correspond to larger effective displacements in the original feature space. Despite this increased sensitivity at extreme noise levels, the robustness observed in the low-noise regime ($\sigma < 0.2$) is sufficient for diffusion-based generation, where the final denoising steps operate in this range.

\FloatBarrier
\section{Normalizing the KL Penalty Across Latent Dimensionalities}
\label{sec:kl-scaling}

When comparing VAE configurations with different latent dimensionalities, the standard $\beta$-VAE objective introduces a confound: the KL divergence term scales with the number of latent dimensions, causing models with larger latent spaces to experience disproportionately stronger regularization. We describe a simple normalization scheme that ensures fair comparison across latent sizes.

\paragraph{Background.} The $\beta$-VAE objective decomposes into a reconstruction term and a KL regularization term:
\begin{equation}
    \mathcal{L} = \mathcal{L}_{\text{recon}} + \beta \cdot D_{\mathrm{KL}}\!\left(q_\phi(\mathbf{z}|\mathbf{x}) \,\|\, p(\mathbf{z})\right),
\end{equation}
where $q_\phi(\mathbf{z}|\mathbf{x})$ is the approximate posterior and $p(\mathbf{z})$ is the prior.

For a factorial Gaussian posterior $q_\phi(\mathbf{z}|\mathbf{x}) = \mathcal{N}(\boldsymbol{\mu}, \mathrm{diag}(\boldsymbol{\sigma}^2))$ and a standard normal prior $p(\mathbf{z}) = \mathcal{N}(\mathbf{0}, \mathbf{I})$, the KL divergence admits a closed-form solution that decomposes as a sum over the $D$ latent dimensions:
\begin{equation}
    D_{\mathrm{KL}} = \sum_{i=1}^{D} \frac{1}{2}\left(\sigma_i^2 + \mu_i^2 - 1 - \log \sigma_i^2\right).
\end{equation}

\paragraph{The scaling problem.} Let $\bar{d}_{\mathrm{KL}}$ denote the average per-dimension KL contribution. Under the assumption that per-dimension statistics are approximately constant across configurations, the total KL scales linearly with the latent dimensionality:
\begin{equation}
    D_{\mathrm{KL}} \approx D \cdot \bar{d}_{\mathrm{KL}}.
\end{equation}

Consider two models with latent dimensions $D^{(1)}$ and $D^{(2)}$ trained with the same $\beta$. Their respective losses are:
\begin{align}
    \mathcal{L}^{(1)} &= \mathcal{L}_{\text{recon}}^{(1)} + \beta \cdot D^{(1)} \cdot \bar{d}_{\mathrm{KL}}, \\
    \mathcal{L}^{(2)} &= \mathcal{L}_{\text{recon}}^{(2)} + \beta \cdot D^{(2)} \cdot \bar{d}_{\mathrm{KL}}.
\end{align}
The effective regularization strength is $\beta \cdot D$, which differs between models. This asymmetry penalizes larger latent spaces more heavily, potentially degrading their reconstruction quality relative to smaller configurations.

\paragraph{Normalization.} To ensure that the per-dimension regularization pressure remains constant across configurations, we require:
\begin{equation}
    \beta^{(1)} \cdot D^{(1)} = \beta^{(2)} \cdot D^{(2)} = \text{const},
\end{equation}
which implies $\beta \propto 1/D$. Given a reference coefficient $\beta_{\text{ref}}$ calibrated at dimensionality $D_{\text{ref}}$, the appropriately scaled coefficient for a model with latent dimension $D$ is:
\begin{equation}
    \beta = \beta_{\text{ref}} \cdot \frac{D_{\text{ref}}}{D}.
\end{equation}

In our experiments, we use $D_{\text{ref}} = 512$ (corresponding to 32 tokens $\times$ 16 features) with $\beta_{\text{ref}} = 10^{-6}$, following the hyperparameter choice from \citet{kimDemocratizingTexttoImageMasked2025}. This normalization ensures that all configurations experience equivalent per-dimension KL pressure, enabling fair comparison of reconstruction quality across different latent sizes. \Cref{tab:kl-weights} lists the resulting $\beta$ values for the latent dimensionalities used in our experiments.

\begin{table}[h]
\centering
\caption{KL weight ($\beta$) values for each latent dimensionality, normalized to maintain consistent per-dimension regularization pressure.}
\label{tab:kl-weights}
\begin{tabular}{cc}
\toprule
\textbf{Latent dim} ($T \times d$) & \textbf{$\beta$} \\
\midrule
256 & $2 \times 10^{-6}$ \\
512 & $1 \times 10^{-6}$ \\
1024 & $5 \times 10^{-7}$ \\
2048 & $2.5 \times 10^{-7}$ \\
4096 & $1.25 \times 10^{-7}$ \\
\bottomrule
\end{tabular}
\end{table}

\FloatBarrier
\section{FlatDINO Training Details}
\label{sec:flatdino-details}

\Cref{tab:flatdino-hparams} summarizes the architectural and optimization hyperparameters for FlatDINO training. Following \citet{kimDemocratizingTexttoImageMasked2025}, we use a ViT-B encoder and ViT-L decoder. The encoder processes 256 DINOv2 patch embeddings along with learnable register tokens, producing latent representations in the registers. The decoder maps these compressed tokens back to 256 spatial features matching the original DINOv2 output.

\begin{table}[h]
\centering
\caption{FlatDINO training hyperparameters.}
\label{tab:flatdino-hparams}
\begin{small}
\begin{tabular}{ll}
\toprule
\textbf{Parameter} & \textbf{Value} \\
\midrule
\multicolumn{2}{l}{\textit{Architecture}} \\
\midrule
Encoder & ViT-B (768 dim, 12 layers, 12 heads) \\
Decoder & ViT-L (1024 dim, 24 layers, 16 heads) \\
Activation & SwiGLU \\
Normalization & LayerNorm \\
\midrule
\multicolumn{2}{l}{\textit{Optimization}} \\
\midrule
Optimizer & AdamW ($\beta_1 = 0.9$, $\beta_2 = 0.999$) \\
Weight decay & 0.02 (2D weights only) \\
Batch size & 512 \\
Precision & bfloat16 compute, float32 params \\
\midrule
\multicolumn{2}{l}{\textit{Learning rate}} \\
\midrule
Peak LR & $10^{-4}$ \\
Min LR & $10^{-8}$ \\
Warmup LR & $10^{-6}$ \\
Schedule & Warmup-Stable-Decay (WSD) \\
\midrule
\multicolumn{2}{l}{\textit{Loss}} \\
\midrule
Reconstruction & Mean squared error \\
KL weight $\beta$ & Normalized by latent dim (\cref{sec:kl-scaling}) \\
\bottomrule
\end{tabular}
\end{small}
\end{table}

\paragraph{Learning rate schedule.} We employ a Warmup-Stable-Decay (WSD) schedule: linear warmup from $10^{-6}$ to $10^{-4}$ over 5 epochs, followed by a stable phase at peak learning rate, and finally cosine decay to $10^{-8}$. WSD enables resource-efficient experimentation, as checkpoints from the stable phase can be resumed for additional training by appending a decay phase. \Cref{tab:wsd-schedule} shows the schedule configurations used in this work.

\begin{table}[h]
\centering
\caption{WSD learning rate schedule configurations.}
\label{tab:wsd-schedule}
\begin{small}
\begin{tabular}{cccc}
\toprule
\textbf{Total epochs} & \textbf{Warmup} & \textbf{Stable} & \textbf{Decay} \\
\midrule
50 & 5 & 40 & 5 \\
150 & 5 & 123 & 22 \\
\bottomrule
\end{tabular}
\end{small}
\end{table}

Checkpoints are selected based on minimum validation reconstruction loss.

\FloatBarrier
\section{Token Ablation: Additional Configurations}
\label{sec:token-ablation-configs}

We extend the token ablation analysis from the main text to additional FlatDINO configurations.

\begin{figure}[h]
\centering
\begin{subfigure}[b]{0.49\textwidth}
    \includegraphics[width=\textwidth]{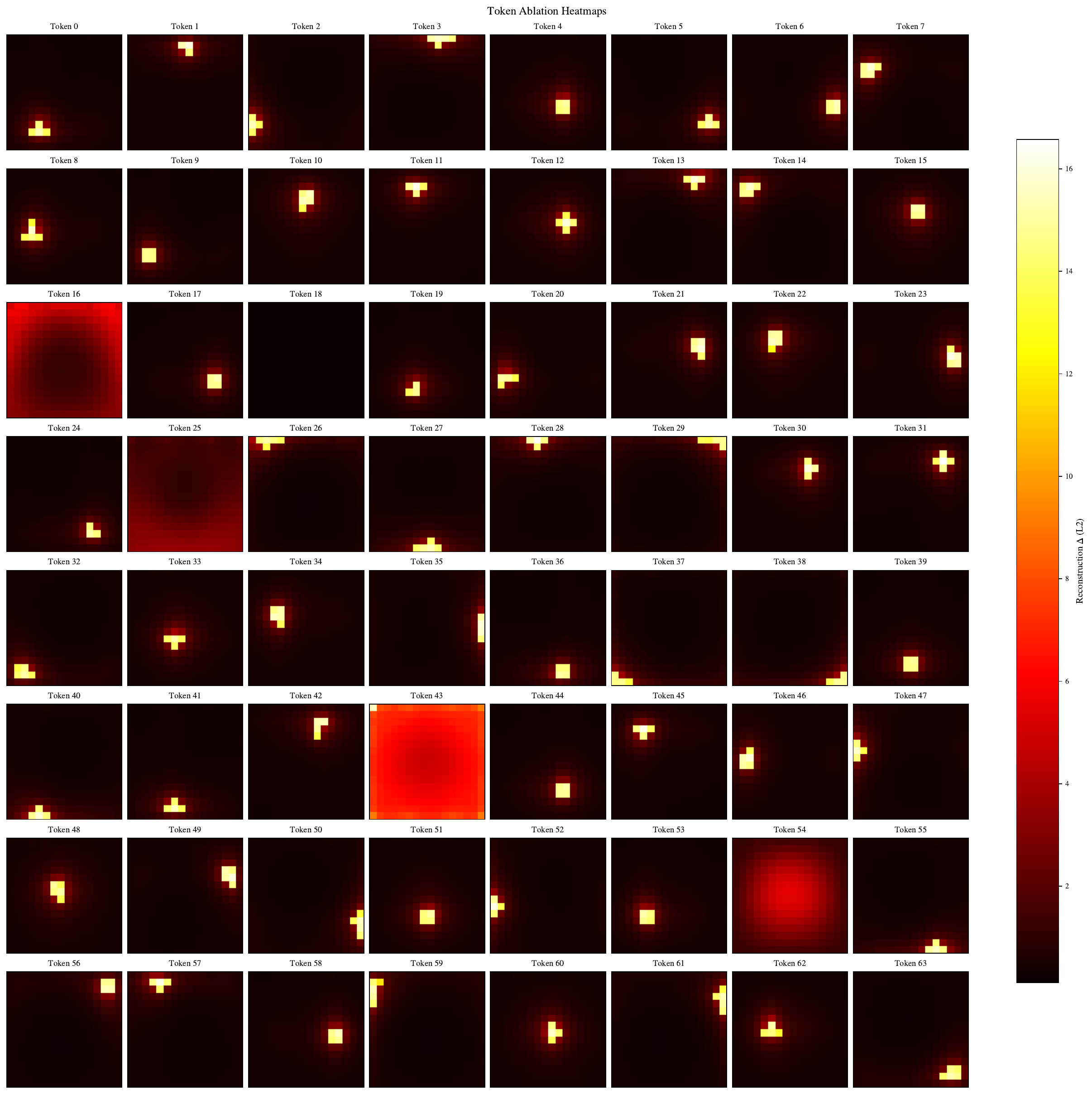}
    \caption{64-token model (64$\times$64).}
\end{subfigure}
\hfill
\begin{subfigure}[b]{0.49\textwidth}
    \includegraphics[width=\textwidth]{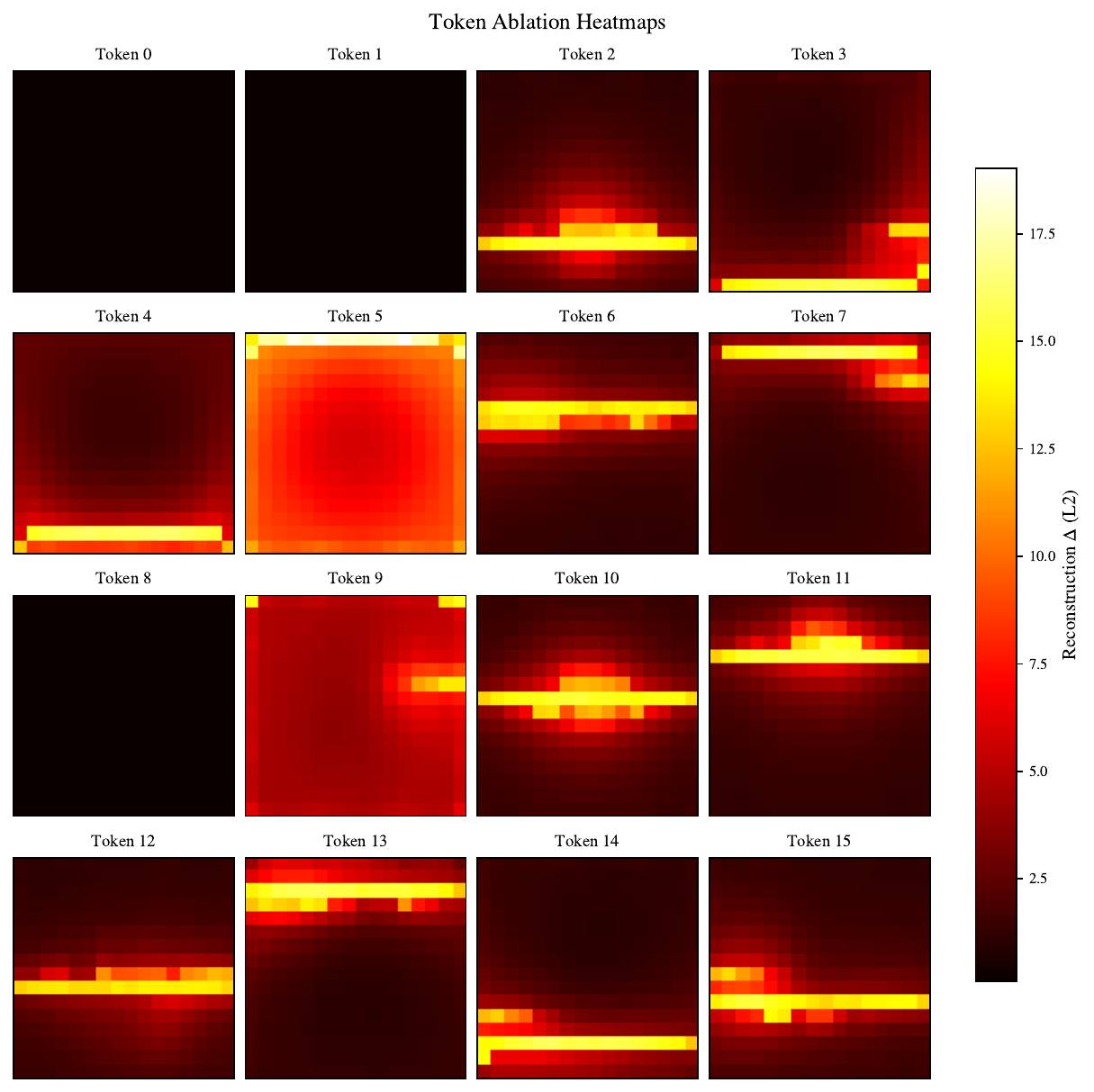}
    \caption{16-token model (16$\times$128).}
\end{subfigure}
\caption{Token ablation heatmaps for varying token counts. Each subplot shows the mean reconstruction error when that token is zeroed out, averaged over 10,000 ImageNet validation images. The 64-token model (left) learns localized blob-like receptive fields similar to the 32-token model, while the 16-token model (right) reverts to encoding horizontal stripes---a qualitatively different spatial organization.}
\label{fig:token-ablation-appendix}
\end{figure}

\paragraph{Varying token count.} \Cref{fig:token-ablation-appendix} compares the 64-token and 16-token models. The 64-token model (left) exhibits spatial organization very similar to the 32-token case. Each token focuses on a small, localized blob of the image, though these blobs are smaller than in the 32-token configuration since the image is partitioned across more tokens. Tokens 16, 25, 43, and 54 are exceptions, displaying global influence on the reconstructed patches rather than spatially localized receptive fields.

The 16-token model (right) shows qualitatively different behavior: rather than encoding localized blobs, most tokens learn to represent entire horizontal stripes of the image. This consistency between 32 and 64 tokens---in contrast to the qualitative shift observed with 16 tokens---suggests that the blob-based encoding strategy is stable above a certain token count threshold.

\begin{figure}[h]
\centering
\begin{minipage}[t]{0.48\textwidth}
\centering
\includegraphics[width=\textwidth]{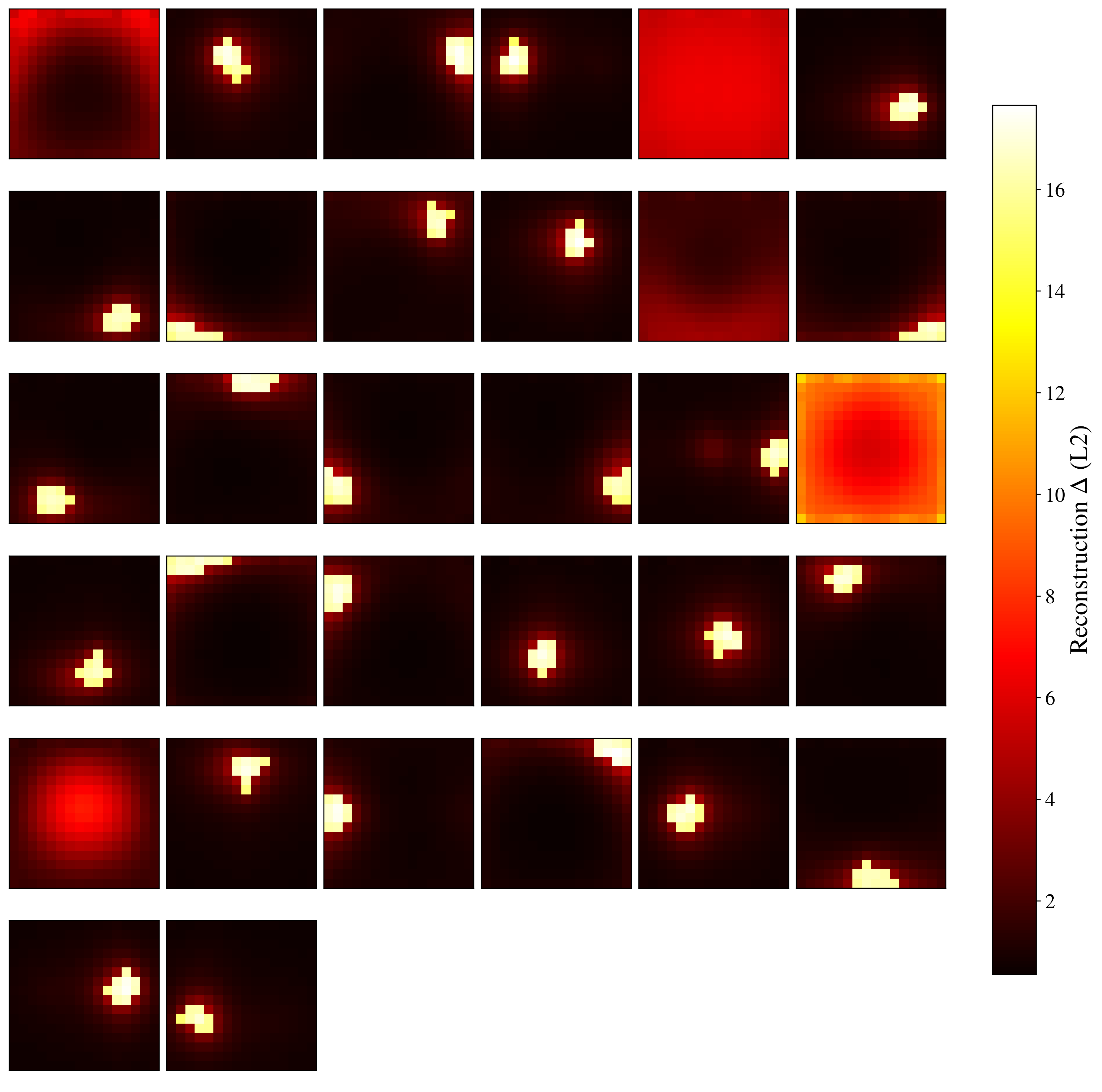}
\captionof{figure}{Token ablation heatmaps for the 32$\times$64 FlatDINO configuration. Five tokens (0, 4, 10, 17, 24) show diffuse global influence.}
\label{fig:token-ablation-32x64}
\end{minipage}
\hfill
\begin{minipage}[t]{0.48\textwidth}
\centering
\includegraphics[width=\textwidth]{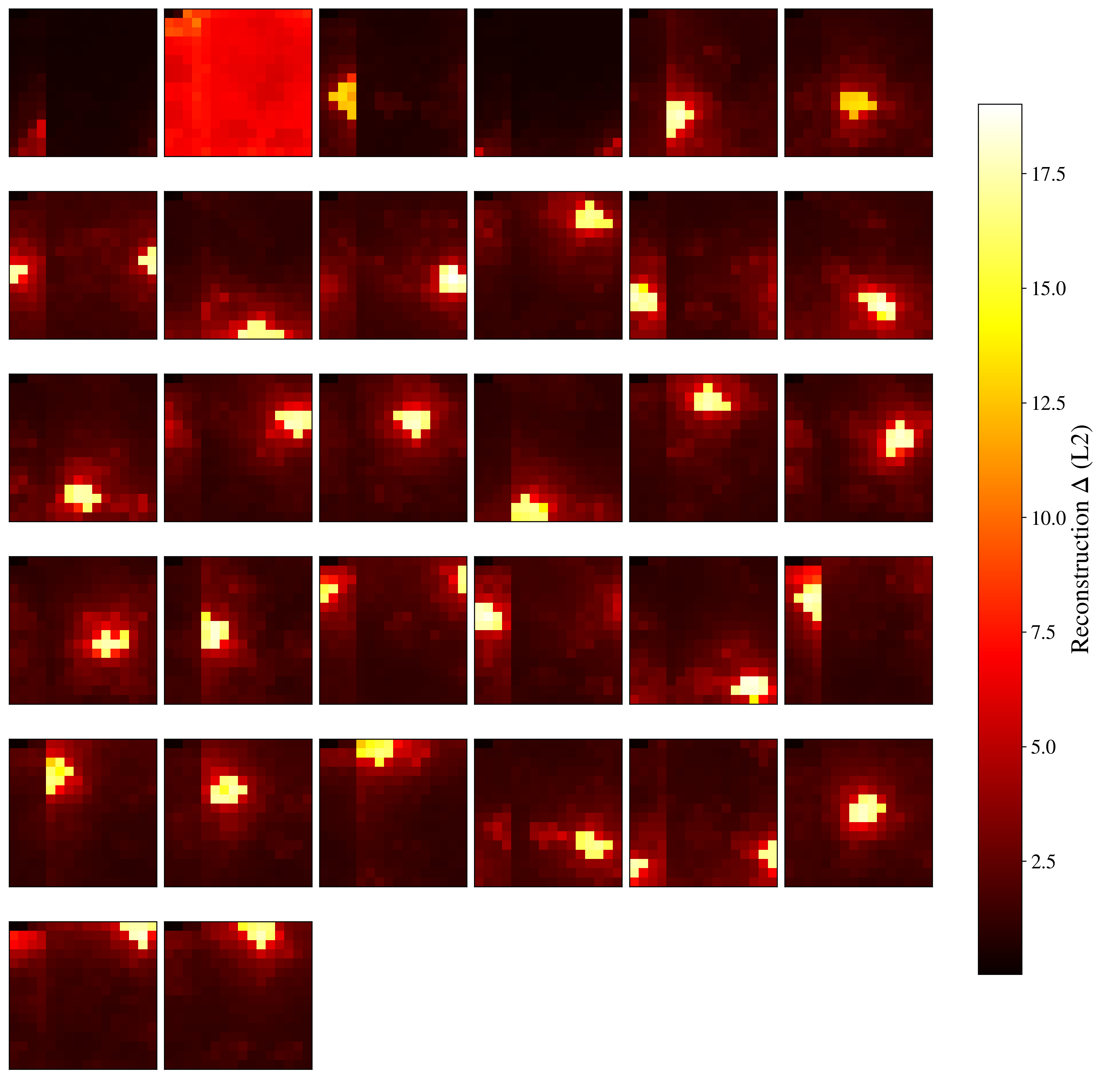}
\captionof{figure}{Token ablation with 4 additional register tokens. The model still dedicates latent tokens to encoding background regions.}
\label{fig:token-ablation-registers}
\end{minipage}
\end{figure}

\paragraph{Varying feature dimension.} \Cref{fig:token-ablation-32x64} shows the token ablation for the 32$\times$64 configuration. While most tokens learn localized receptive fields similar to the 32$\times$128 model shown in the main text, five tokens (0, 4, 10, 17, 24) exhibit diffuse, image-wide influence rather than spatially localized responses, suggesting they encode global rather than local information.

\paragraph{Effect of register tokens.} \citet{darcetVisionTransformersNeed2024} observed that ViT backbones develop high-norm tokens in low-information background regions, and proposed adding register tokens to absorb this global information. We tested whether this approach could eliminate the background-encoding token in FlatDINO by adding 4 registers to both the encoder and decoder. As shown in \cref{fig:token-ablation-registers}, the phenomenon persists: the model still dedicates tokens to encoding background regions even with explicit register tokens available.

\FloatBarrier
\section{CFG Hyperparameter Sweep}
\label{sec:cfg-sweep}

\begin{figure}[h]
\centering
\includegraphics[width=0.49\textwidth]{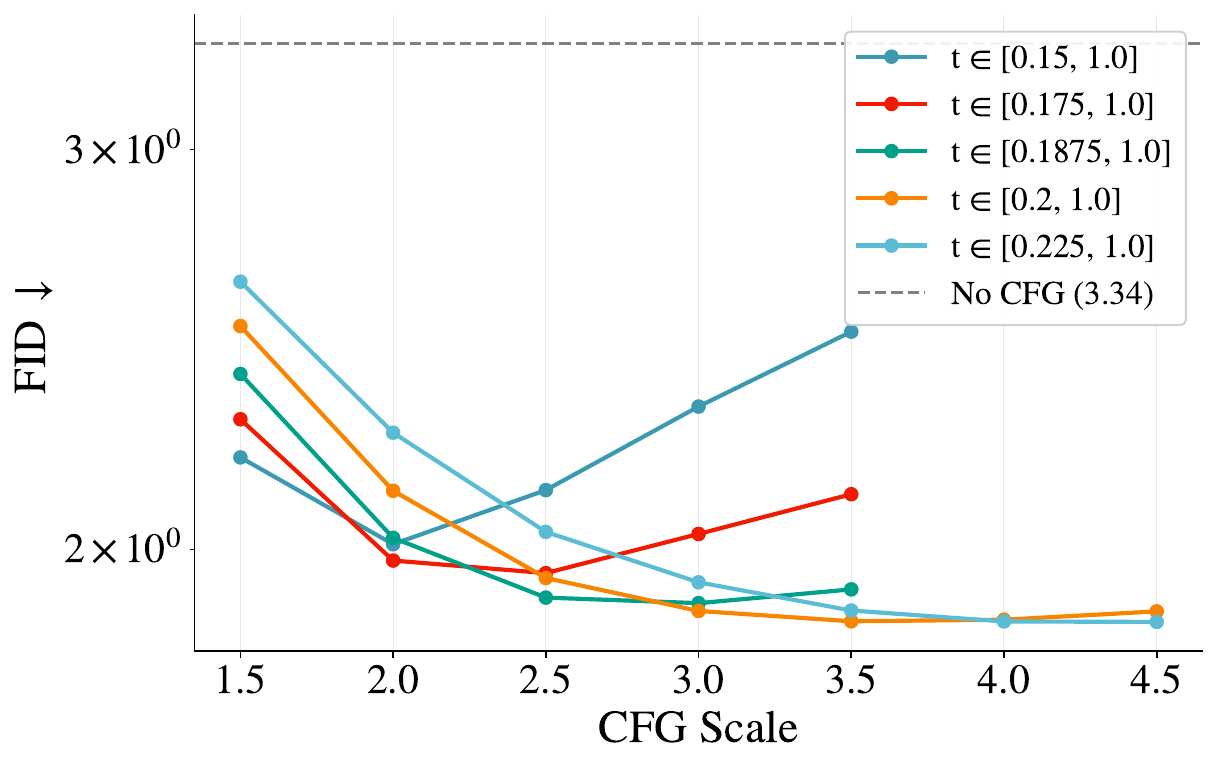}
\caption{CFG hyperparameter sweep showing gFID as a function of guidance weight and starting time interval. Lower gFID indicates better generation quality.}
\label{fig:cfg-sweep}
\end{figure}

Following \citet{kynkaanniemiApplyingGuidanceLimited2024}, we apply classifier-free guidance only during a limited time interval rather than throughout the full diffusion process. To select the guidance weight and starting time, we perform a grid search over CFG weights and interval starting points, evaluating gFID on a subset of generated samples. \Cref{fig:cfg-sweep} shows the results of this sweep. We select a guidance weight of 4.5 applied during $t \in [0.225, 1.0]$, which achieves the best trade-off between sample quality and diversity.

\FloatBarrier
\section{Inference Performance Profiling}
\label{sec:inference-profiling}

\begin{figure}[h]
\centering
\includegraphics[width=0.49\textwidth]{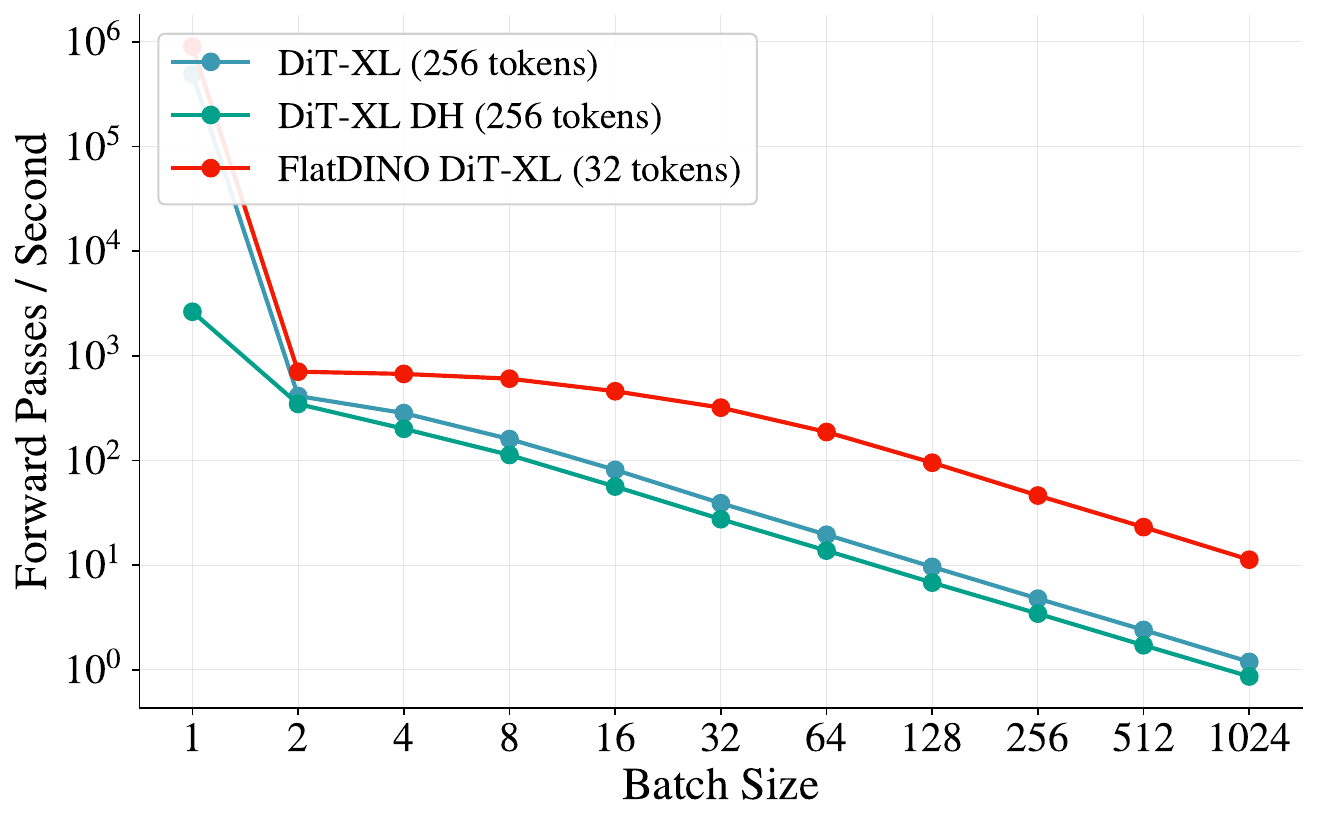}
\caption{Inference throughput on NVIDIA H100 NVL. FlatDINO (32 tokens) versus RAE (256 tokens).}
\label{fig:perf-h100}
\end{figure}

\begin{figure}[h]
\centering
\includegraphics[width=0.49\textwidth]{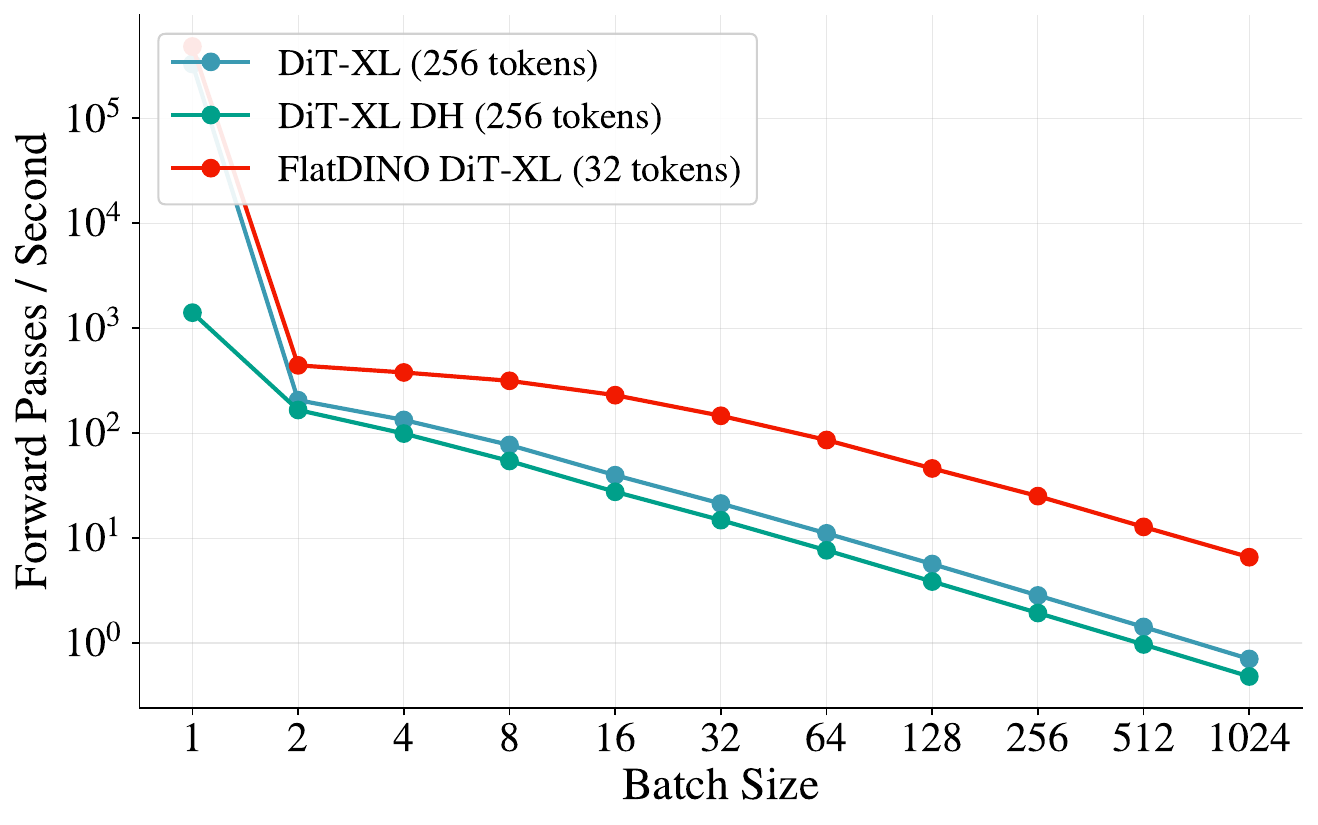}
\caption{Inference throughput on NVIDIA A100 80GB PCIe. FlatDINO (32 tokens) versus RAE (256 tokens).}
\label{fig:perf-a100}
\end{figure}

\begin{figure}[h]
\centering
\includegraphics[width=0.49\textwidth]{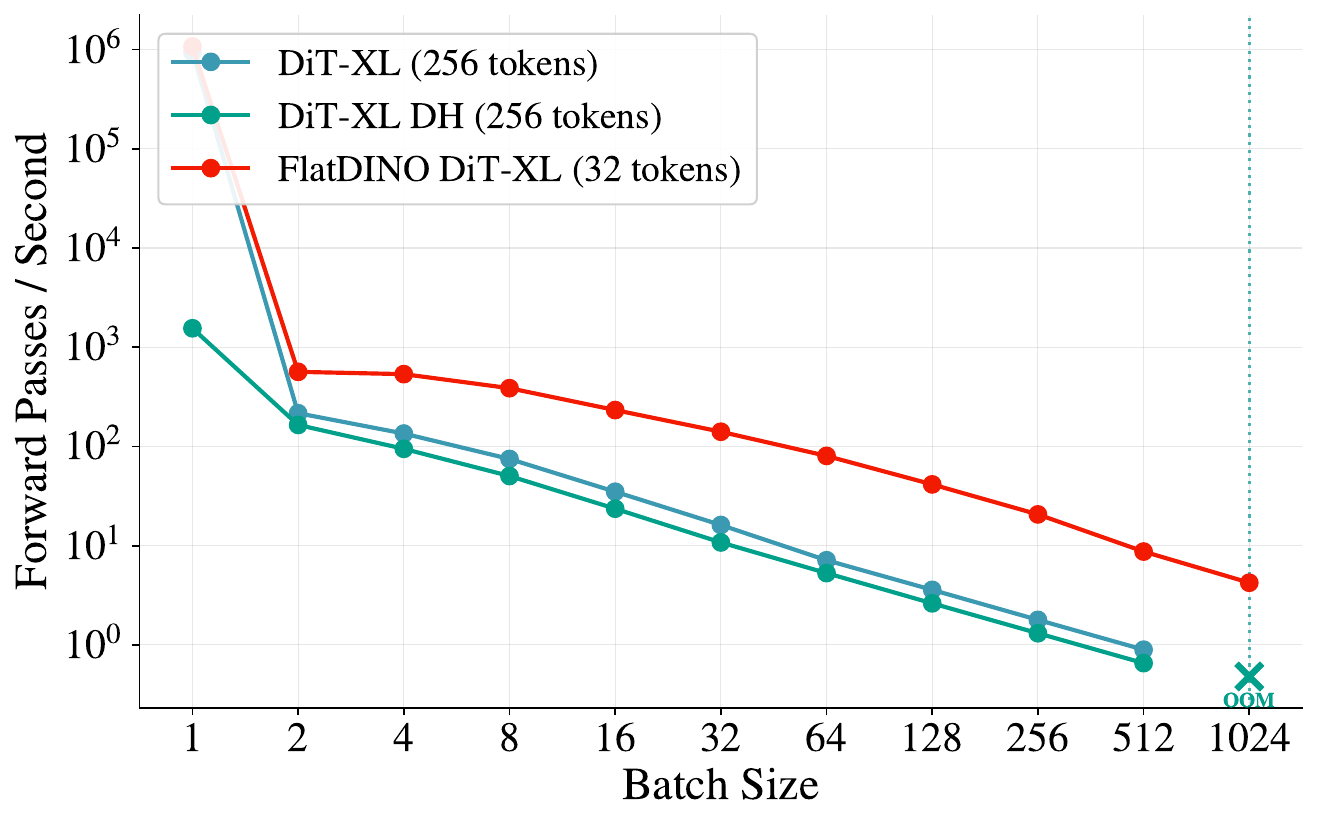}
\caption{Inference throughput on NVIDIA RTX 4090. FlatDINO (32 tokens) versus RAE (256 tokens).}
\label{fig:perf-rtx4090}
\end{figure}

To validate the theoretical efficiency gains of FlatDINO, we benchmark the inference throughput of DiT-XL operating on FlatDINO latents (32 tokens) versus RAE latents (256 tokens) across three GPU architectures. \Cref{fig:perf-h100,fig:perf-a100,fig:perf-rtx4090} show the number of forward passes per second as a function of batch size.

Even at small batch sizes, FlatDINO demonstrates measurable performance improvements over the RAE baseline. As batch size increases, the throughput gap widens and approaches the theoretical $8\times$ speedup predicted by the token count reduction. This behavior reflects the improved hardware utilization at larger batch sizes, where the reduced memory footprint of FlatDINO's shorter sequences allows for more efficient parallelization.

\FloatBarrier
\section{FLOPs Comparison}
\label{sec:flops}

We derive the computational cost of a transformer layer with parameter-matched SwiGLU \citep{shazeer2020gluvariantsimprovetransformer} and use it to compare the FLOPs of diffusion models operating on RAE (256 tokens) versus FlatDINO (32 tokens).

\subsection{FLOPs per Transformer Layer}

Consider a transformer with batch size $B$, sequence length $S$, and hidden dimension $D$. We count one multiply-add as one FLOP and report per-sample values (i.e., $B = 1$); other conventions rescale all entries equally.

\paragraph{Self-attention.} The attention mechanism involves:
\begin{itemize}
    \item QKV projections: $3 \times B \times S \times D^2$ FLOPs
    \item Attention scores ($QK^\top$): $B \times S^2 \times D$ FLOPs
    \item Attention-weighted values: $B \times S^2 \times D$ FLOPs
    \item Output projection: $B \times S \times D^2$ FLOPs
\end{itemize}
Total attention FLOPs: $4BSD^2 + 2BS^2D$.

\paragraph{SwiGLU FFN.} For parameter matching with a standard FFN (hidden dimension $4D$), SwiGLU uses hidden dimension $H = 8D/3$. The SwiGLU computation involves:
\begin{itemize}
    \item Gate projection: $B \times S \times D \times H$ FLOPs
    \item Value projection: $B \times S \times D \times H$ FLOPs
    \item Down projection: $B \times S \times H \times D$ FLOPs
\end{itemize}
With $H = 8D/3$, total FFN FLOPs: $3 \times BSD \times \frac{8D}{3} = 8BSD^2$.

\paragraph{Total per layer.}
\begin{equation}
    \text{FLOPs}_{\text{layer}} = 12BSD^2 + 2BS^2D = 2BSD(6D + S)
\end{equation}

The linear term ($12BSD^2$) dominates when $D > S/6$; the quadratic attention term ($2BS^2D$) dominates when $S > 6D$.

\subsection{DiT-L and DiT-XL Comparison}

DiT-L uses hidden dimension $D = 1024$ and $L = 24$ layers; DiT-XL uses $D = 1152$ and $L = 28$ layers; $\text{DiT}^{\text{DH}}$-XL adds 2 additional layers with $D = 2048$ for the decoupled head. Since $6D > S$ for all configurations, the linear term dominates.

\begin{table}[h]
\centering
\caption{FLOPs comparison for DiT-L, DiT-XL, and $\text{DiT}^{\text{DH}}$-XL operating on RAE versus FlatDINO latent spaces.}
\label{tab:flops}
\begin{small}
\begin{tabular}{llccccc}
\toprule
\textbf{Model} & \textbf{Latent} & $S$ & $D$ & \textbf{Linear} ($12SD^2$) & \textbf{Attention} ($2S^2D$) & \textbf{Total} \\
\midrule
\multirow{2}{*}{DiT-L (24L)} & RAE & 256 & 1024 & 3.22 GFLOPs & 0.13 GFLOPs & 80.5 GFLOPs \\
 & FlatDINO & 32 & 1024 & 0.40 GFLOPs & 0.002 GFLOPs & 9.7 GFLOPs \\
\midrule
\multirow{2}{*}{DiT-XL (28L)} & RAE & 256 & 1152 & 4.08 GFLOPs & 0.15 GFLOPs & 118.4 GFLOPs \\
 & FlatDINO & 32 & 1152 & 0.51 GFLOPs & 0.002 GFLOPs & 14.3 GFLOPs \\
\midrule
\multirow{2}{*}{$\text{DiT}^{\text{DH}}$-XL (28L+2L)} & RAE & 256 & 1152/2048 & 4.08/12.88 GFLOPs & 0.15/0.27 GFLOPs & 144.7 GFLOPs \\
 & FlatDINO & 32 & 1152/2048 & 0.51/1.61 GFLOPs & 0.002/0.004 GFLOPs & 17.5 GFLOPs \\
\bottomrule
\end{tabular}
\end{small}
\end{table}

FlatDINO achieves an $\mathbf{8.3\times}$ reduction in FLOPs for DiT-L, DiT-XL, and $\text{DiT}^{\text{DH}}$-XL. The reduction is approximately $8\times$ (the ratio of sequence lengths) because the linear term dominates. The quadratic attention term contributes only $\sim$4\% of total FLOPs for RAE, so the 64$\times$ reduction in attention cost has limited impact on overall computation.

\subsection{Training FLOPs with Encoding Overhead}

The forward pass comparison above considers only the DiT. During training, we must also account for:
\begin{enumerate}
    \item \textbf{Backward pass}: Computing gradients requires approximately $2\times$ the forward FLOPs, so a full training step costs $\sim 3\times$ forward FLOPs.
    \item \textbf{Encoding overhead}: Both RAE and FlatDINO require a DINOv2 forward pass; FlatDINO additionally requires the FlatDINO encoder forward pass.
\end{enumerate}

\paragraph{Backward pass FLOPs.} Using the $3\times$ forward approximation:

\begin{table}[h]
\centering
\caption{DiT forward + backward FLOPs per training step.}
\label{tab:backward-flops}
\begin{small}
\begin{tabular}{llccc}
\toprule
\textbf{Model} & \textbf{Latent} & \textbf{Forward} & \textbf{Backward} & \textbf{Total} \\
\midrule
\multirow{2}{*}{DiT-L} & RAE & 80.5 GFLOPs & 161.0 GFLOPs & 241.5 GFLOPs \\
 & FlatDINO & 9.7 GFLOPs & 19.4 GFLOPs & 29.1 GFLOPs \\
\midrule
\multirow{2}{*}{DiT-XL} & RAE & 118.4 GFLOPs & 236.8 GFLOPs & 355.2 GFLOPs \\
 & FlatDINO & 14.3 GFLOPs & 28.6 GFLOPs & 42.9 GFLOPs \\
\midrule
\multirow{2}{*}{$\text{DiT}^{\text{DH}}$-XL} & RAE & 144.7 GFLOPs & 289.4 GFLOPs & 434.1 GFLOPs \\
 & FlatDINO & 17.5 GFLOPs & 35.0 GFLOPs & 52.5 GFLOPs \\
\bottomrule
\end{tabular}
\end{small}
\end{table}

\paragraph{Encoding overhead.} Both RAE and FlatDINO operate on DINOv2 features, so both require the DINOv2 forward pass. FlatDINO additionally requires its encoder. We compute the FLOPs using the same formula:

\begin{itemize}
    \item \textbf{DINOv2 ViT-B}: $D = 768$, $L = 12$, $S = 261$ (256 patches + CLS + 4 registers).
    \begin{equation}
        \text{FLOPs} = 12 \times 2 \times 261 \times 768 \times (6 \times 768 + 261) = 23.4 \text{ GFLOPs}
    \end{equation}
    \item \textbf{FlatDINO encoder ViT-B}: $D = 768$, $L = 12$, $S = 288$ (256 DINO patches + 32 register tokens).
    \begin{equation}
        \text{FLOPs} = 12 \times 2 \times 288 \times 768 \times (6 \times 768 + 288) = 26.0 \text{ GFLOPs}
    \end{equation}
\end{itemize}

\paragraph{Full training step comparison.}

\begin{table}[h]
\centering
\caption{Total FLOPs per training step including encoding overhead.}
\label{tab:training-flops}
\begin{small}
\begin{tabular}{llcccc}
\toprule
\textbf{Model} & \textbf{Latent} & \textbf{Encoding} & \textbf{DiT Train} & \textbf{Total} & \textbf{Reduction} \\
\midrule
\multirow{2}{*}{DiT-L} & RAE & 23.4 GFLOPs & 241.5 GFLOPs & 264.9 GFLOPs & \multirow{2}{*}{$\mathbf{3.4\times}$} \\
 & FlatDINO & 49.4 GFLOPs & 29.1 GFLOPs & 78.5 GFLOPs & \\
\midrule
\multirow{2}{*}{DiT-XL} & RAE & 23.4 GFLOPs & 355.2 GFLOPs & 378.6 GFLOPs & \multirow{2}{*}{$\mathbf{4.1\times}$} \\
 & FlatDINO & 49.4 GFLOPs & 42.9 GFLOPs & 92.3 GFLOPs & \\
\midrule
\multirow{2}{*}{$\text{DiT}^{\text{DH}}$-XL} & RAE & 23.4 GFLOPs & 434.1 GFLOPs & 457.5 GFLOPs & \multirow{2}{*}{$\mathbf{4.5\times}$} \\
 & FlatDINO & 49.4 GFLOPs & 52.5 GFLOPs & 101.9 GFLOPs & \\
\bottomrule
\end{tabular}
\end{small}
\end{table}

Both methods share the DINOv2 encoding cost (23.4 GFLOPs), but FlatDINO adds an additional encoder forward pass (26.0 GFLOPs). Despite this overhead, training on FlatDINO latents requires substantially fewer FLOPs. The $\mathbf{3.4\times}$ reduction for DiT-L, $\mathbf{4.1\times}$ for DiT-XL, and $\mathbf{4.5\times}$ for $\text{DiT}^{\text{DH}}$-XL demonstrate that the 8$\times$ reduction in DiT FLOPs more than compensates for the FlatDINO encoder overhead. The reduction factor improves with larger DiT models, as the fixed encoding cost becomes a smaller fraction of total computation.

\FloatBarrier
\section{k-NN Evaluation}
\label{sec:knn}

We evaluate the quality of the learned representations using $k$-nearest neighbor ($k$-NN) classification, following the protocol described in \citet{caronEmergingPropertiesSelfSupervised2021}, across several image recognition benchmarks: CIFAR-10/100 \citep{Krizhevsky09learningmultiple}, Caltech-101 \citep{fei2007learning}, Oxford Flowers \citep{Nilsback08}, Oxford Pets \citep{parkhi2012cat}, Food101 \citep{bossard14}, and DTD \citep{cimpoi14describing}. For both DINOv2 patch embeddings and FlatDINO tokens, we apply average pooling followed by feature normalization before computing $k$-NN.

As shown in \cref{tab:knn}, compressing only the token count while preserving the full feature dimension (32$\times$768) retains---and even slightly improves---the discriminative properties of DINOv2 patches, with ImageNet k-NN accuracy increasing from 74.3\% to 77.2\%. However, further compressing along the feature dimension degrades performance: accuracy drops to 65.1\% for 32$\times$128 and 46.4\% for 32$\times$64. This suggests that while spatial redundancy in the patch grid can be eliminated without losing semantic information, the feature dimension encodes discriminative content that is harder to compress losslessly.

\begin{table}[h]
\caption{k-NN (Top-1 (\%)) performance evaluation of DINOv2 and FlatDINO on fine-grained benchmarks.}
\label{tab:knn}
\centering
\begin{small}
\begin{tabular}{l|c|ccccccc}
\toprule
Feature & ImageNet-1k & CIFAR-10 & CIFAR-100 & Caltech-101 & Flowers & Pets & Food-101 & DTD \\
\midrule
DINOv2 CLS & \textbf{81.92} & \underline{97.92} & \underline{87.56} & 85.89 & \textbf{99.69} & \textbf{94.52} & \textbf{88.65} & \underline{76.75} \\
DINOv2 Patches & 74.34 & 97.69 & 87.13 & \textbf{89.46} & 99.41 & 76.06 & 82.52 & \textbf{77.12} \\
\midrule
FlatDINO $32x768$ & \underline{77.20} & \textbf{98.08} & \textbf{88.94} & \underline{88.19} & \underline{99.63} & \underline{87.16} & \underline{84.23} & 76.65 \\
FlatDINO $32x128$ & 65.11 & 89.86 & 71.01 & 85.68 & 95.67 & 69.07 & 78.33 & 76.60 \\
FlatDINO $32x64$ & 46.36 & 80.77 & 60.15 & 81.77 & 74.70 & 47.10 & 59.81 & 72.13 \\
\bottomrule
\end{tabular}
\end{small}
\end{table}

\FloatBarrier
\section{Token Inversion}
\label{sec:inversion-details}

DINOv2 with registers produces a CLS token and four register tokens in addition to the patch embeddings. We investigated whether these tokens encode spatial structure that could be leveraged for compression, using the Deep Image Prior (DIP) inversion method \citep{ulyanovDeepImagePrior2020, tumanyanSplicingViTFeatures2022}.

Following \citet{tumanyanSplicingViTFeatures2022}, we initialize a frozen tensor $z \in \mathbb{R}^{H\times W\times D}$ with white noise and feed it into a trainable U-Net. The U-Net is optimized to minimize the squared error between the encoder output of its generated image and a given target token. Optimization proceeds with $H,W$ matching the image size and $D=32$. We use AdamW with gradient clipping at norm 10.0 and a learning rate of 0.01. Noise regularization is applied with stage-dependent variance: $\sigma_1=10.0$ for the first 10,000 steps, $\sigma_2=2.0$ for the subsequent 5,000 iterations, and $\sigma_3=0.5$ for the final 5,000 iterations.

\begin{figure}[h]
\centering
\begin{subfigure}[b]{0.3\textwidth}
    \includegraphics[width=\textwidth]{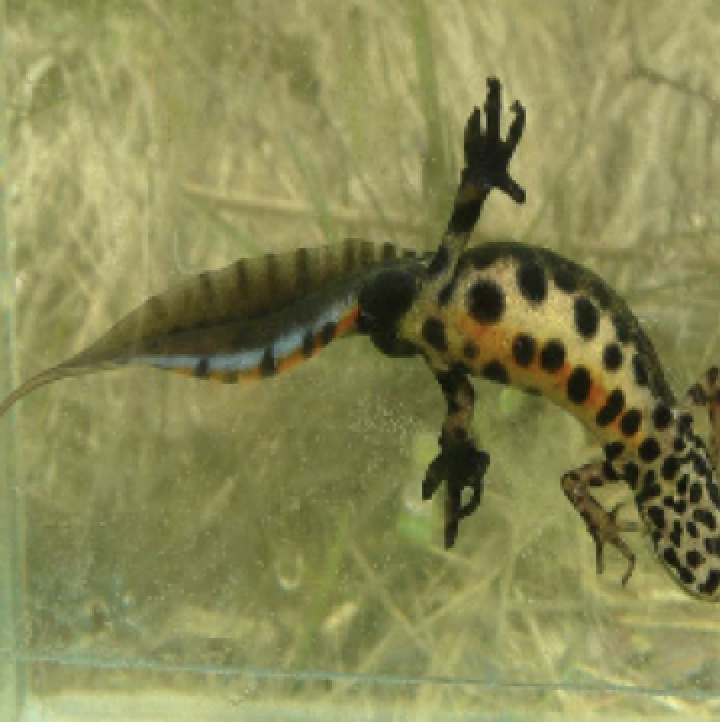}
    \caption{Original image.}
\end{subfigure}
\begin{subfigure}[b]{0.3\textwidth}
    \includegraphics[width=\textwidth]{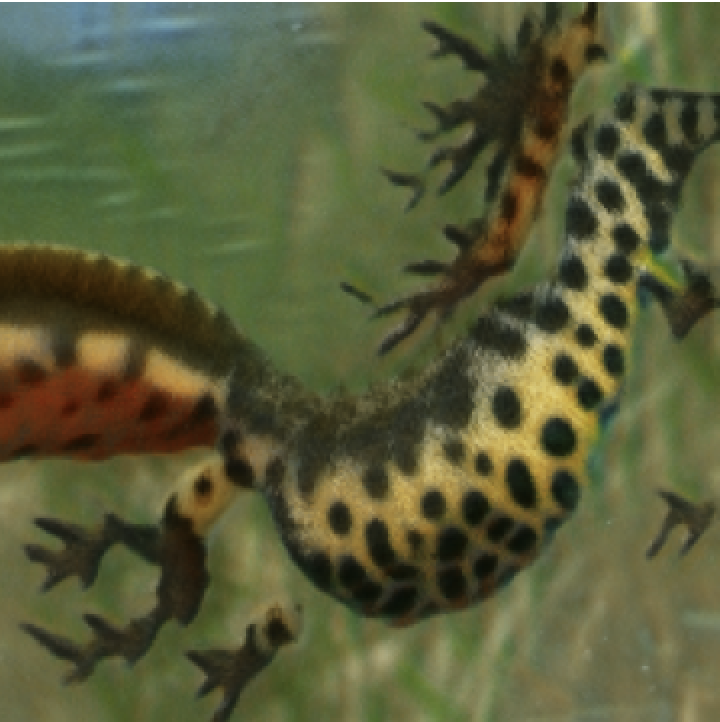}
    \caption{DINOv2 CLS.}
\end{subfigure}
\begin{subfigure}[b]{0.3\textwidth}
    \includegraphics[width=\textwidth]{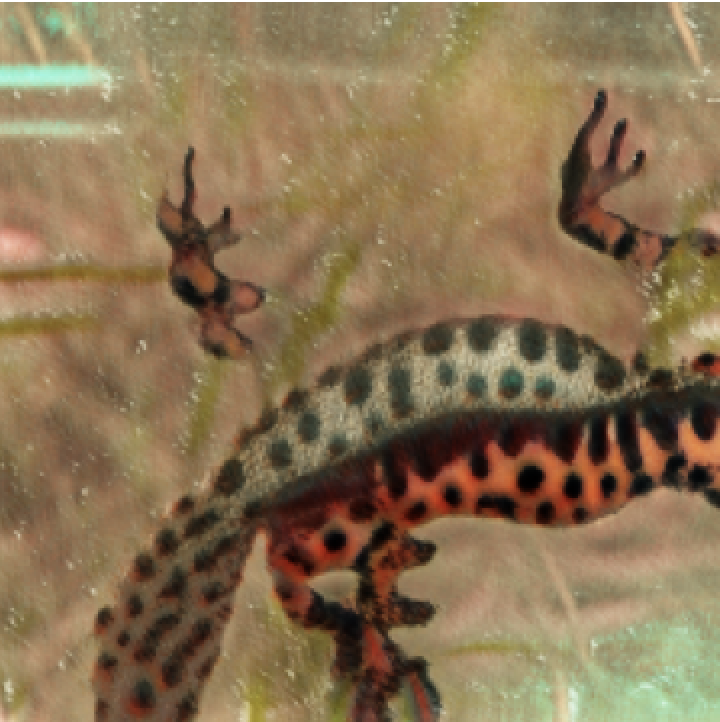}
    \caption{DINOv2 Registers.}
\end{subfigure}
\caption{Inversion of DINOv2's CLS and register tokens. (b) The CLS token captures texture and semantics but misses the scene layout. (c) The register tokens recover more structure yet still misplace objects and orientations.}
\label{fig:inversion_dino}
\end{figure}

As shown in \cref{fig:inversion_dino}, the CLS token captures texture and high-level semantics but lacks spatial layout information---the dog's pose and position are not recovered. This is expected, as the CLS token is trained to be invariant to the augmentations applied during DINO's self-supervised training (cropping, flipping, color jittering, etc.). The register tokens preserve slightly more structure but still misplace objects and orientations. This absence of reliable spatial information in the CLS and register tokens motivated our approach of learning new compressed tokens from the spatially-organized patch embeddings rather than repurposing DINOv2's existing global tokens.

\end{document}